\documentclass[journal]{IEEEtran}
\usepackage{blindtext}
\usepackage{graphicx}

\usepackage{algorithm}
\usepackage{algpseudocode}
\usepackage{amsmath}

\usepackage{amssymb}

\usepackage{tikz}

\usepackage{ifpdf}
 \ifpdf
 \else
 \fi
\usepackage{lipsum,multicol}
\usepackage{cite}
\usepackage{array}
\usepackage{amssymb,epsfig,latexsym,epsf,color}
\usepackage{algorithm, algorithmicx, algpseudocode}
\usepackage{graphicx,amsmath,amssymb,latexsym}
\usepackage{subfigure}
\usepackage{paralist}
\usepackage{bm}
\usepackage{times}
\usepackage{multirow}
\usepackage{color}
\usepackage{balance}
\usepackage{verbatim}
\usepackage{amsthm}
\usepackage{mathabx}

\ifCLASSINFOpdf

\else

\fi

\usepackage{url}

\usepackage{amssymb}
\usepackage{amsmath}
\usepackage{booktabs}

\begin{document}

\title{Federated Learning-Enabled Hybrid Language Models for Communication-Efficient Token Transmission}
%Latency Minimization on Hybrid Federated and Centralized Learning Over Wireless Communication Networks}
%% 
\author{Faranaksadat~Solat,\;\IEEEmembership{Member,~IEEE,} Joohyung~Lee,\;\IEEEmembership{Senior Member,~IEEE,}\;Mohamed~Seif, \;\IEEEmembership{Member,~IEEE,} Dusit~Niyato,\;\IEEEmembership{Fellow,~IEEE}, and H. Vincent~Poor,\;~\IEEEmembership{Life Fellow,~IEEE}

\thanks{Faranaksadat~Solat and Joohyung~Lee are with the Department of Computing, Gachon Univ., Seongnam 13120, Rep. of Korea (e-mail: faranak1995@gachon.ac.kr; j17.lee@gachon.ac.kr).}

\thanks{Dusit~Niyato is with the School of Computer Engineering, Nanyang
Technological University, Singapore (e-mail: dniyato@ntu.edu.sg).}

\thanks{Mohamed~Seif and H. Vincent Poor with the Department of Electrical and Computer Engineering, Princeton University, Princeton, New Jersey, 08544, US (email: mseif@princeton.edu; poor@princeton.edu).}

\thanks{Corresponding author: Joohyung Lee}}

\markboth{}%
{Shell \MakeLowercase{\textit{et al.}}: Bare Demo of IEEEtran.cls for IEEE Journals}

\maketitle

\begin{abstract}
Hybrid Language Models (HLMs) are inference-time architectures that combine the low-latency efficiency of Small Language Models (SLMs) on clients (edge devices) with the high accuracy of Large Language Models (LLMs) in centralized servers. Unlike traditional end-to-end LLM inference, HLMs aim to reduce latency and communication by selectively invoking LLMs only when the local SLM's predictions are uncertain---that is, when the model exhibits low confidence or high entropy in its token-level probability distribution. However, when the SLM encounters ambiguous or low-confidence predictions during inference, it must offload token-level probability distributions to the LLM for refinement. This frequent offloading leads to substantial communication overhead, particularly in bandwidth-constrained environments. To address this challenge, we propose FedHLM, a communication-efficient HLM framework that integrates uncertainty-aware inference with Federated Learning (FL). The key innovation lies in collaboratively learning token-level uncertainty thresholds that determine when SLM predictions require LLM assistance. Instead of relying on static or hand-tuned thresholds, FedHLM uses FL to enable distributed threshold optimization across clients while preserving data privacy. Additionally, embedding-based token representations are employed to facilitate semantic similarity comparisons during Peer-to-Peer (P2P) resolution, allowing clients to reuse tokens inferred by similar peers without efficiently involving the LLM. Moreover, we propose hierarchical model aggregation as a strategy to reduce redundant token transmissions. At the edge server level, client updates are aggregated to refine local routing policies, while global coordination across clusters further synchronizes decision boundaries. This layered approach ensures that repeated uncertainty patterns are captured and resolved locally, significantly reducing unnecessary LLM queries. Extensive simulations on large-scale news classification tasks demonstrate that FedHLM achieves over 95\% reduction in LLM transmissions with negligible accuracy loss, highlighting its potential for scalable and efficient edge-Artificial Intelligence (AI) deployment.
\end{abstract}


\IEEEpeerreviewmaketitle
\section{Introduction} \label{section.introduction}

\IEEEPARstart{T}{he} remarkable progress of Large Language Models (LLMs) has enabled state-of-the-art performance across a wide range of Natural Language Processing (NLP) tasks. Despite their effectiveness, the deployment of LLMs on resource-constrained clients (edge devices) remains impractical due to high computational, memory, and energy demands \cite{Solat2024_SFL, Lee2024_FL}. To address these limitations, Hybrid Language Models (HLMs) have emerged as a promising solution \cite{shao2025division}, combining Small Language Models (SLMs) deployed locally on clients with LLMs operating in centralized cloud servers. This cooperative inference paradigm allows lightweight local processing while offloading complex or ambiguous tasks to powerful cloud infrastructure.

\subsection{Communication Bottlenecks in HLM}

A major bottleneck in HLM frameworks lies in the frequent transmission of token-level probability distributions from SLMs to LLMs. These transmissions, triggered by low-confidence tokens, incur significant communication overhead, increase end-to-end inference latency, and degrade scalability in bandwidth-constrained wireless networks \cite{Solat2024_het}. To mitigate this, prior works have proposed token-level routing and uncertainty-aware fallback mechanisms. In token routing schemes \cite{qiao2025token}, systems dynamically decide whether to accept a locally generated token or forward it to the server based on predefined rules, often leveraging semantic cues or entropy-based thresholds. Similarly, uncertainty-aware fallback strategies \cite{Wang2025_uncertainity} estimate the model's prediction confidence—commonly using techniques such as dropout variance or ensemble disagreement—and offload only the uncertain tokens to the cloud for validation or refinement.

While effective to some extent, these methods typically rely on static thresholds or handcrafted heuristics. Such thresholds are difficult to tune manually and often fail to adapt to the heterogeneous nature of edge environments, where user behavior, input complexity, and device resources vary widely. In practice, an overly conservative threshold may result in excessive offloading, while an overly permissive one could reduce accuracy due to premature acceptance of uncertain tokens. This creates a need for a more adaptive and personalized mechanism for threshold selection.

Moreover, most fallback schemes treat the cloud-based LLM as the sole decision authority for uncertain tokens, overlooking the potential for collaboration among similar clients. In many real-world applications—such as smart home assistants or regional Question Answering (QA) systems—clients exhibit semantic overlap in token usage. Ignoring this redundancy leads to inefficient offloading and avoidable LLM queries. PromptFL \cite{Guo2024promptfl} offers an alternative direction by enabling clients to collaboratively learn prompts instead of full models, improving personalization and communication efficiency. However, such approaches are limited to prompt adaptation and do not directly address token-level transmission optimization during inference.

To overcome these challenges, we propose FedHLM, a communication-efficient HLM framework that integrates federated threshold learning with peer-aware token resolution. By collaboratively learning personalized uncertainty thresholds, FedHLM adapts token routing decisions to local client contexts. Furthermore, by leveraging semantic similarity among peer clients for token reuse, FedHLM reduces reliance on LLM queries and enhances scalability in bandwidth-constrained environments.

\subsection{Overview of the Proposed FedHLM Framework}

In this paper, we propose FedHLM, a communication-efficient HLM framework that integrates federated learning (FL) with uncertainty-guided inference. FL is a decentralized learning paradigm in which multiple clients collaboratively train a shared model by transmitting local updates rather than raw data, thereby preserving data privacy and reducing centralized communication overhead \cite{mcmahan2017communication, Lee2024_FL}. This makes FL particularly well-suited for bandwidth-constrained environments with privacy-sensitive data, such as mobile or Internet of Things (IoT) systems.

In the context of HLMs, where each edge client may encounter distinct linguistic patterns or application-specific inputs, static or global thresholds fail to account for personalization needs. FedHLM addresses this by enabling clients to collaboratively learn optimal token-level uncertainty thresholds through a decentralized FL process. These thresholds govern when to accept a local SLM prediction and when to invoke fallback mechanisms such as peer reuse or LLM resolution. Unlike fixed thresholds that are manually tuned or heuristically applied, our learned thresholds adapt to each client's local uncertainty patterns, leading to more context-sensitive routing decisions.

In addition, FedHLM leverages semantic token embeddings to facilitate peer-aware resolution. When a token is flagged as uncertain, the system first searches for semantically similar predictions among peers by comparing token embeddings. This embedding-based comparison enables high-fidelity token reuse across clients without requiring redundant LLM queries or exact token matching, which is particularly beneficial in natural language contexts where tokens with similar meaning may differ in form. By using token-level embeddings rather than discrete IDs, FedHLM achieves more robust and scalable peer collaboration.

To support this collaborative learning, FedHLM adopts a hierarchical FL architecture specifically confined to the threshold optimization process. This structure is not used for full model training or fine-tuning, but instead organizes clients into semantic clusters and coordinates threshold learning across both intra-cluster and inter-cluster levels. Each cluster aggregates its members' threshold updates locally, and these are further synchronized globally, ensuring both efficient adaptation and cross-domain generalization. Compared to flat aggregation, this hierarchical scheme reduces uplink communication costs and improves convergence speed by limiting global synchronization to representative cluster-level statistics.

This federated architecture enables scalable and adaptive token filtering across heterogeneous clients without requiring access to raw user data, while maintaining communication efficiency and inference accuracy.

\begin{table*}[htbp]
\centering
\caption{Comparison of Key Papers and Our FedHLM Framework}
\label{tab:comparison}
\begin{tabular}{|p{2.5cm}|p{4cm}|p{4.2cm}|p{4.2cm}|} 
\hline
\textbf{Feature} & \textbf{Token Communications Paper} & \textbf{Uncertainty-Aware HLM Paper} & \textbf{Our FedHLM Framework} \\
\hline
\textbf{Focus} & Cross-modal token-based communication & Hybrid inference with uncertainty-aware token selection & FL-driven optimization for token transmission \\
\hline
\textbf{Communication Efficiency} & Uses token compression but does not adaptively minimize transmissions & Reduces transmissions using fixed uncertainty thresholds & Dynamically minimizes uplink communication using FL \\
\hline
\textbf{Adaptation} & No learning-based adaptation & Uses static transmission thresholds & Uses FL to learn personalized transmission thresholds per device \\
\hline
\textbf{Scalability} & Not designed for multi-device environments & Works for a single-device HLM setup & Scalable across multiple clients using FL \\
\hline
\end{tabular}
\end{table*}

\subsection{Summary of Contributions}

The main contributions of this paper are summarized as follows:

\begin{itemize}
    \item We propose a novel framework, FedHLM, to extend token-based communication by integrating FL for optimizing uncertainty-driven token transmission in HLMs. Unlike centralized approaches, FedHLM enables distributed adaptation of transmission thresholds across clients, significantly reducing communication overhead while maintaining inference accuracy.
    
    \item A key novelty of FedHLM lies in its federated aggregation of token-level uncertainty thresholds. Each client locally optimizes a token transmission boundary by monitoring when the SLM rejects a prediction, and these local thresholds are aggregated using Federated Averaging (FedAvg) to form a global threshold policy. This collaborative process adapts routing behavior based on real-world inference uncertainty without requiring data centralization. FedHLM specifically balances a trade-off between transmission efficiency and semantic fidelity—where transmission efficiency refers to minimizing the number of tokens sent to peers or the cloud, and semantic fidelity refers to preserving the original meaning or intent of the input despite early exits or local inference. By learning thresholds that are sensitive to local uncertainty while being informed by global trends, FedHLM ensures that only genuinely ambiguous tokens are escalated, thereby reducing unnecessary transmission while maintaining accurate and semantically faithful responses.

    \item To support this framework, we develop a formal mathematical model for optimizing token transmission. The formulation incorporates uncertainty estimation, probabilistic decision-making, and FL-based gradient updates, providing a principled strategy to minimize redundant transmissions while preserving model accuracy.

    \item Finally, a scalable federated token processing architecture is introduced, moving beyond centralized semantic tokenization. The design includes a hierarchical FL structure that clusters clients based on semantic similarity to support efficient threshold aggregation across local and global levels. This enables personalized token selection and adaptive routing with minimal server reliance. Furthermore, we implement the proposed system as a modular Kubeflow-based pipeline, demonstrating practical feasibility and compatibility with real-world deployment in wireless edge environments.

    \item Extensive simulations demonstrate that FedHLM reduces LLM token transmissions by over 95\% compared to baseline methods (Rand-HLM and U-HLM), while maintaining high accuracy and ensuring scalability across federated clients and communication rounds.
\end{itemize}

\subsection{Paper Organization}

The remainder of this paper is organized as follows. Section 2 reviews related work on hybrid language models, FL for NLP, and uncertainty-aware inference. Section 3 presents the system model and formal problem formulation, including the federated threshold optimization. Section 4 details the proposed FedHLM framework and its hierarchical architecture. Section 5 provides convergence insights, communication trade-off analysis, and the role of token entropy in caching efficiency. Section 6 discusses the experimental setup and presents evaluation results. Finally, Section 7 concludes the paper and outlines future research directions.

\section{RELATED WORKS} \label{sec:relatedworks}
In this section, we review the existing literature relevant to our study, focusing on three key areas: HLMs, FL for NLP, and uncertainty-aware inference in communication-efficient learning systems.

\subsection{Hybrid Language Models}
HLMs combine the computational efficiency of SLMs at the edge with the expressive power of LLMs at the server. These models aim to reduce dependency on cloud-based processing while maintaining high inference quality. Recent studies \cite{gu2023efficiently, chen2019speculative, shao2025division, Hao2024_HLM, Hao2024_HLM_cite} have explored different hybrid architectures where clients generate \textit{preliminary} token predictions, and cloud-based LLMs refine or correct them. However, many existing approaches rely on \textit{frequent} transmissions of token distributions, leading to increased communication overhead. Speculative decoding \cite{chen2019speculative} and token-level caching techniques \cite{mcmahan2017communication} have been proposed to alleviate some of these challenges, but they typically use fixed thresholds or heuristics, lacking adaptability to diverse client behaviors and local input characteristics. 

Our work builds upon these HLM paradigms by integrating a federated threshold adaptation mechanism that allows clients to collaboratively learn personalized token transmission boundaries. This enables more efficient and context-aware inference without relying on centralized tuning or hand-crafted rules.

\subsection{FL for NLP}
FL has been widely applied in NLP tasks, enabling decentralized model training without sharing raw data \cite{GarciaBernal2020, mcmahan2017communication, konecny2017federated, Yuan2022_dec}. Traditional FL in NLP has been used to fine-tune large-scale transformer models across multiple devices while preserving data privacy \cite{mcmahan2017communication, Lin2021_Fed_NLP, Cai2023_FL, Cai2023_NLP, Liu2021_FL}. However, most FL approaches focus on model training rather than optimizing token-level inference. Recent works, such as TITANIC \cite{su2024titanic}, have explored FL for personalized inference, adapting model parameters based on local user interactions, but they do not specifically address communication-efficient hybrid inference. In our work, we extend FL beyond model training by using it to \textit{dynamically} optimize token transmission decisions, reducing communication costs while maintaining high-quality inference.

\subsection{Uncertainty-Aware Inference for Efficient Communication}
Uncertainty-aware inference techniques have been introduced in deep learning to selectively prioritize data transmission or computation \cite{gal2016dropout, zhang2020uncertainty, Mobiny2021, Labach2019}. In NLP, uncertainty estimation is commonly used for active learning \cite{settles2009active, Yang2016_active, Zhu2010_active, Nguyen2022} and confidence-based filtering \cite{blatz2004confidence, Sharma2025}. Some recent studies \cite{zhang2020uncertainty} have integrated uncertainty measures into adaptive transmission systems, where only high-uncertainty predictions are transmitted for correction. In speculative inference settings \cite{chen2019speculative}, models attempt to reduce computation by predicting token sequences that are likely to be accepted by a more complex model. However, these techniques often rely on \textit{predefined} thresholds and do not leverage federated optimization for personalized and decentralized uncertainty handling. Our approach introduces an FL-driven adaptive uncertainty thresholding mechanism that enables each client to learn a transmission policy tailored to its local input characteristics. While it does not explicitly optimize for network or device heterogeneity, the reduction in redundant uplink transmissions indirectly contributes to improved communication efficiency under constrained conditions.

While previous works have explored hybrid inference, FL-based NLP models, and uncertainty-aware transmission, they remain largely \textit{independent} fields. Our work bridges these areas by introducing a FL-enabled hybrid language model (FedHLM) that optimizes token transmissions based on dynamically learned uncertainty thresholds. This integration significantly reduces communication overhead while preserving model accuracy, making it suitable for real-time inference in edge-cloud environments. To further clarify the distinctions between existing studies and our proposed approach, Table~\ref{tab:comparison} provides a comparative summary across key features. While prior works have made progress in token compression and uncertainty-based inference, they typically rely on fixed thresholds or centralized decision-making. In contrast, FedHLM integrates uncertainty estimation with federated optimization and peer-to-peer (P2P) collaboration, enabling adaptive, scalable, and communication-efficient token resolution in edge-cloud settings.

The next section presents the system model and problem formulation, detailing the mathematical foundations of our proposed FedHLM framework.

\begin{table}[htbp]
\centering
\caption{Summary of Notation}
\label{tab:notations}
\begin{tabular}{|c|p{6.5cm}|}
\hline
\textbf{Symbol} & \textbf{Description} \\
\hline
$\beta_t^k$ & Rejection probability from LLM for token $x_t^k$ \\
\hline
$C_k$ & Local token cache at client $k$ \\
\hline
$C_{\text{LLM}}$ & Communication cost for token resolution via LLM \\
\hline
$C_{\text{P2P}}$ & Communication cost for P2P token resolution \\
\hline
$C_{\text{uplink}}$ & Uplink communication cost per token \\
\hline
$d_k$ & Resampled token from temperature-adjusted SLM sampling \\
\hline
$\delta_t^k$ & Routing decision (approximated by sigmoid) for token $x_t^k$ at client $k$, time $t$ \\
\hline
$\mathbf{e}_t^k$ & Embedding of token predicted at client $k$, time $t$ \\
\hline
$\bar{\mathbf{e}}_t^{(c)}$ & Centroid embedding of tokens in cluster $c$ at time $t$\\
\hline
$H(S)$ & Cache hit ratio as a function of cache size $S$ \\
\hline
$H_k$ & Token entropy at client $k$ \\
\hline
$K$ & Number of clients \\
\hline
$M$ & Number of edge clusters \\
\hline
$n_k$ & Number of transmitted tokens at client $k$ \\
\hline
$T$ & Total number of FL rounds \\
\hline
$V$ & Vocabulary size \\
\hline
$p_{\text{hit}}$ & Probability of successful peer resolution \\
\hline
$\eta$ & Learning rate for threshold updates \\
\hline
$\gamma$ & Temperature scaling parameter for sigmoid approximation \\
\hline
$\lambda$ & Regularization parameter in local loss function \\
\hline
$\sigma(z)$ & Sigmoid function $\sigma(z) = \frac{1}{1 + e^{-z}}$ \\
\hline
$\text{sim}_t^k$ & Cosine similarity of token embeddings for client $k$ at time $t$ \\
\hline
$\theta$ & Cosine similarity threshold for peer consensus \\
\hline
$\tau_{\text{LLM}}$ & Latency of cloud LLM inference \\
\hline
$\tau_{\text{SLM}}$ & Latency of local SLM inference \\
\hline
$\tau_{\text{uplink}}$ & Latency of uplink token transmission \\
\hline
$\mathbf{y}^{(t)}$ & Token probability distribution from LLM at time $t$ \\
\hline
$\hat{u}_{\text{th},c}$ & Cluster-level aggregated threshold for cluster $c$ \\
\hline
$\mathcal{L}_k(u_{\text{th},k})$ & Local loss function at client $k$ for optimizing threshold $u_{\text{th},k}$ \\
\hline
$\frac{\partial \mathcal{L}_k}{\partial u_{\text{th},k}}$ & Gradient of the local loss w.r.t. $u_{\text{th},k}$ \\
\hline
$u_t^k$ & Uncertainty score for token $x_t^k$ at client $k$ \\
\hline
$u_{\text{th}}$ & Global aggregated uncertainty threshold (via FedAvg) \\
\hline
$u_{\text{th},k}$ & Local uncertainty threshold at client $k$ \\
\hline
$\mathbf{x}^{(t)}$ & Token probability distribution from SLM at time $t$ \\
\hline
$x_i^{(t)}$ & The $i$-th sampled token from SLM distribution at time $t$ \\
\hline
$x_t^k$ & Token predicted by client $k$'s SLM at timestep $t$ \\
\hline
\end{tabular}
\end{table}

\section{Proposed FedHLM Training \& Inference Framework} \label{sec:model}
In this section, we define the system model for the proposed FedHLM. The proposed system follows a hierarchical architecture comprising clients running SLMs, intermediate edge servers coordinating clustered clients, and a central cloud server hosting the LLM. The goal is to optimize token transmission between the SLMs and the LLM by leveraging FL to minimize communication overhead while maintaining inference accuracy. Table~\ref{tab:notations} summarizes the key notation used throughout the system model and optimization formulations.

Furthermore, to support scalable and personalized aggregation, clients are pre-clustered based on semantic similarity. In FedHLM, semantic similarity refers to the degree of alignment between clients' token-level prediction distributions or embedding representations. It captures how similarly clients interpret and generate token outputs, particularly under domain-specific or repeated inputs. In practice, this is estimated using a lightweight classifier trained on token embedding statistics or output entropy profiles. Grouping clients with high semantic similarity ensures that federated updates are aggregated among peers with similar inference behavior, thereby improving convergence and preserving personalization.

\begin{figure*}[ht!]
\centering
\setkeys{Gin}{width=\linewidth}
    \begin{subfigure}
        \centering        \includegraphics[width=500pt,keepaspectratio]{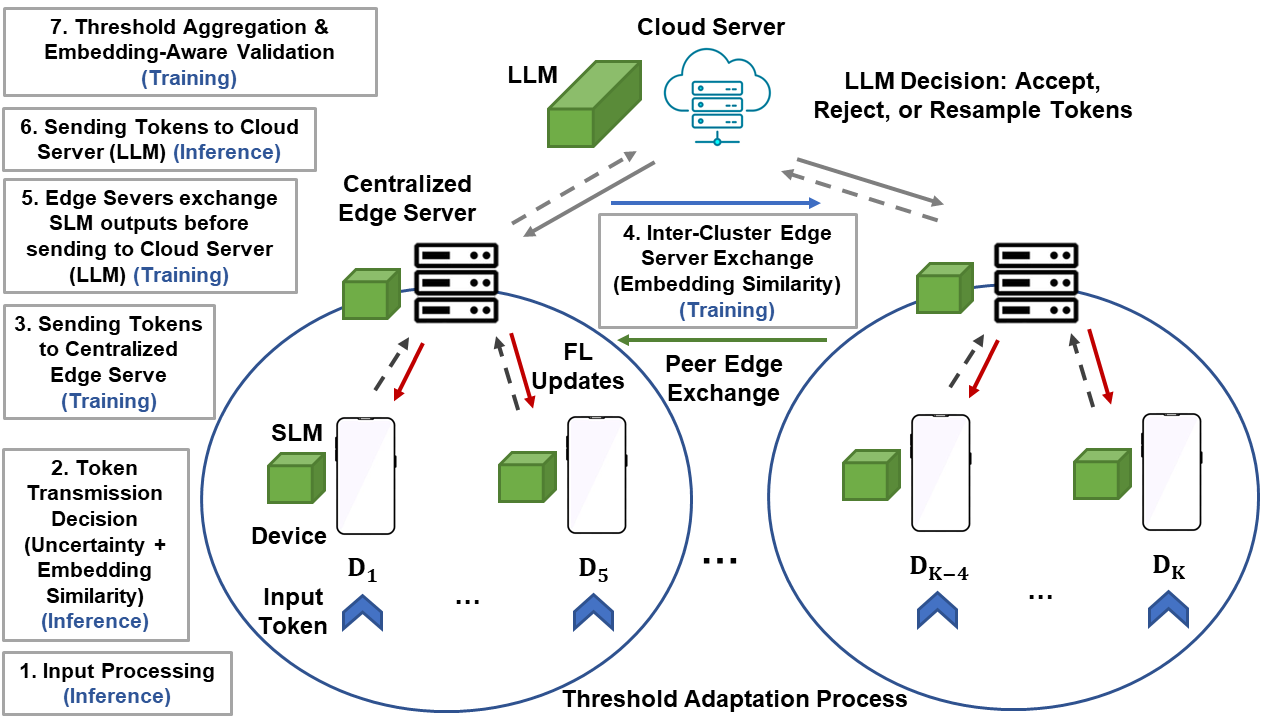}
        \caption{System architecture of the proposed FedHLM framework for communication-efficient hybrid language inference. The client layer performs lightweight inference using SLMs and applies token-level uncertainty thresholds to determine whether to process tokens locally or transmit them for collaborative resolution. Clients are pre-clustered using a classifier-based similarity assessment inspired by SnapCFL~\cite{Cheng_snapcfl}, in which a binary classifier is trained to measure semantic distribution similarity across devices. These clusters form the basis for edge-level FL and token threshold adaptation. Within each cluster, uncertain tokens are routed to a centralized edge server that supports peer-aware model exchange and inter-cluster token validation using embedding similarity. Tokens that remain unresolved are forwarded to a cloud-based LLM for final decisions, including acceptance, rejection, or resampling. The figure annotates each step as part of either the inference or training phase, helping to distinguish runtime behavior from periodic federated threshold updates. This hierarchical client–edge–cloud architecture enables scalable, distributed token inference in resource-constrained wireless environments.}
        \label{fig:system_model}
    \end{subfigure}
    \hfill
\end{figure*}

\subsection{Proposed System Architecture}
The FedHLM framework adopts a three-tier hierarchical architecture comprising clients, edge servers, and a central cloud server, as illustrated in Fig.~\ref{fig:system_model}. This figure provides a high-level overview of the system’s operational flow and the interaction between its key components. These components are defined as follows:

\begin{itemize}
    \item Clients: Represented at the bottom layer of Fig.~\ref{fig:system_model}, each client hosts a lightweight SLM for local inference. To reduce communication with the cloud-based LLM, each device selectively transmits only uncertain tokens. Clients also participate in intra-cluster FL and engage in peer-aware token resolution with other devices in the same cluster.

    \item Edge Servers: As shown in the middle layer of Fig.~\ref{fig:system_model}, clients are pre-clustered based on semantic similarity using a classifier-based strategy inspired by SnapCFL. Each cluster is coordinated by an edge server, which aggregates model updates from clients, facilitates peer-aware model exchange, and performs inter-cluster token validation. The edge server reduces unnecessary communication with the cloud by resolving many uncertain tokens locally.

    \item Cloud Server (LLM + FL Coordinator): As shown at the top of Fig.~\ref{fig:system_model}, the central server hosts the LLM, responsible for final refinement of unresolved tokens. It also acts as the FL coordinator by aggregating threshold update vectors (e.g., confidence margins or transmission policies) from all clusters using FedAvg and distributing the updated global strategy back to edge servers.
\end{itemize}

\begin{figure}[htbp]
    \centering
    \includegraphics[width=0.47\textwidth]{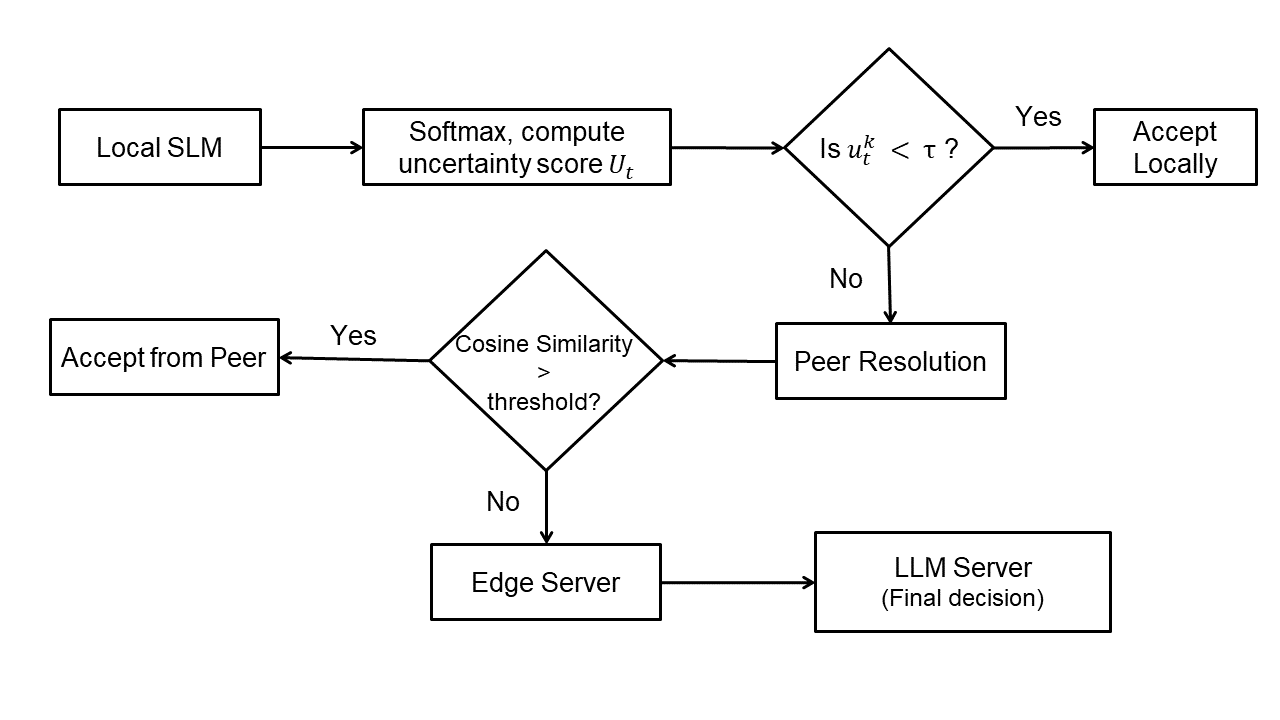} % Make sure to save the image with this name
    \caption{Token routing flow in FedHLM. Each client begins by performing inference with its local SLM and computing the uncertainty score for each token. If the uncertainty is below a learned threshold $\tau$, the token is accepted locally. Otherwise, the token is evaluated through peer-aware resolution within the client cluster using cosine similarity between token embeddings. If the token is still unresolved, it is forwarded to the edge server and ultimately to the LLM for final decision-making. This multi-stage routing mechanism enables communication-efficient inference by escalating only high-uncertainty tokens.}
    \label{fig:flowchart}
\end{figure}

\textit{Fig.~\ref{fig:flowchart}} illustrates this process, showing how uncertainty scores govern the multi-step routing flow across the FedHLM architecture.

\subsection{System Model and Token Transmission}
\label{subsec:system_model}

The proposed FedHLM framework, illustrated in Fig.~\ref{fig:system_model}, follows a hierarchical and communication-efficient pipeline for hybrid language inference and federated optimization. The process unfolds in the following steps:

\begin{itemize}
    \item \textbf{Step 1: Local SLM Inference.} Each client performs lightweight inference using its locally deployed SLM, generating token predictions along with an associated uncertainty score. The SLM computes a probability distribution over the vocabulary $\mathcal{V}$ via a softmax layer, where $P(y_t \mid x)$ denotes the token probability at time step $t$.

    \item \textbf{Step 2: Local Filtering.} A token is defined as the smallest prediction unit (e.g., subword or wordpiece), represented by an embedding vector $\mathbf{e}_t \in \mathbb{R}^d$. To assess prediction confidence, each token is assigned an uncertainty score $U_t$ computed as the entropy of its probability distribution:

    \[
    U_t = - \sum_{i=1}^{|\mathcal{V}|} P(y_t = v_i \mid x) \log P(y_t = v_i \mid x),
    \]

    where $v_i \in \mathcal{V}$. If $U_t < \tau$, the locally learned threshold, the token is accepted without incurring transmission cost.

    \item \textbf{Step 3: Escalation to Peer Resolution.} Tokens exceeding the uncertainty threshold undergo peer-aware resolution. Clients are pre-clustered using a semantic similarity-based method inspired by SnapCFL \cite{Cheng_snapcfl}. Semantic similarity reflects alignment in clients’ token prediction behavior, approximated using token embedding statistics and entropy profiles.

    \item \textbf{Step 4: Intra-Cluster Exchange.} Within each cluster, clients share embedding vectors of uncertain tokens with peers. Each client computes the cosine similarity between its token and the centroid of peer embeddings. If the similarity exceeds a predefined threshold, the token is accepted locally based on peer consensus.

    \item \textbf{Step 5: Inter-Cluster Validation.} Unresolved tokens are forwarded to the edge server. The server optionally performs embedding-based comparison across neighboring clusters to attempt further resolution.

    \item \textbf{Steps 6–7: LLM Decision.} Remaining unresolved tokens are escalated to the cloud-hosted LLM, which issues a final decision: accept, reject, or resample.

    \item \textbf{Federated Threshold Optimization.} In parallel, each client updates its local uncertainty threshold based on token transmission feedback. Using FedAvg, intra-cluster updates are aggregated at the edge server, then globally averaged at the cloud server and redistributed to clients. This enables decentralized but collaborative adaptation of token routing behavior.
\end{itemize}

This multi-stage routing and training strategy ensures that only high-uncertainty tokens are escalated, reducing bandwidth while maintaining inference quality. The integration of semantic filtering and federated threshold learning enables scalable and adaptive deployment across heterogeneous edge environments.

\section{Proposed FedHLM Scheme – Hierarchical Federated Learning for Dynamic Threshold Optimization} \label{sec:proposed_algorithms}

This section formalizes the learning strategy used in the proposed framework to reduce communication with the cloud-based LLM while maintaining inference accuracy. The approach combines uncertainty-aware token routing decisions with a federated threshold adaptation mechanism that enables collaborative and personalized transmission behavior across clients.

\subsection{Token Generation and Rejection Feedback}

At each time step $ t $, a client samples a token $ x_t \in \{1, \dots, |V|\} $ from the SLM’s predicted vocabulary distribution:
\begin{equation}
    x_t \sim \mathbf{x}^{(t)} = [x_1^{(t)}, x_2^{(t)}, \dots, x_{|V|}^{(t)}],
\end{equation}
where \( \mathbf{x}^{(t)} \in \mathbb{R}^{|V|} \) is the softmax output of the SLM at time \( t \), and \( |V| \) is the vocabulary size. Here, \( x_i^{(t)} = P(y_t = v_i \mid x_{<t}) \) represents the probability of predicting token \( v_i \in \mathcal{V} \) at position \( t \), given the input prefix \( x_{<t} \).

The corresponding prediction from the cloud-hosted LLM is given by:
\begin{equation}
    \mathbf{y}^{(t)} = [y_1^{(t)}, y_2^{(t)}, \dots, y_{|V|}^{(t)}],
\end{equation}
where \( \mathbf{y}^{(t)} \in \mathbb{R}^{|V|} \) denotes the LLM’s softmax output distribution at time step \( t \). Each element \( y_i^{(t)} = P_{\text{LLM}}(y_t = v_i \mid x_{<t}) \) represents the probability that the LLM predicts token \( v_i \in \mathcal{V} \), given the input context \( x_{<t} \).

To evaluate the disagreement between the local SLM and the cloud-based LLM, we define a rejection probability for token \( x_t \) as:
\begin{equation}
    \beta_t^k = \max\left(1 - \frac{y^{(t)}[x_t]}{x^{(t)}[x_t]}, 0\right),
    \label{eq:rejection-prob}
\end{equation}
where, \( x^{(t)}[x_t] \) and \( y^{(t)}[x_t] \) denote the softmax probabilities assigned to token \( x_t \) by the SLM and LLM, respectively. Intuitively, \( \beta_t^k \in [0,1] \) quantifies the LLM's lack of agreement with the SLM’s decision: a high value indicates that the LLM assigns a much lower probability to the token than the SLM, suggesting likely rejection.

This formulation captures the confidence gap between models and has been adapted from confidence-based filtering ideas in prior speculative inference and fallback systems (e.g., \cite{chen2019speculative, zhang2020uncertainty}). Note that \( \beta_t^k \) is computed after transmission, based solely on model outputs and is not dependent on the threshold \( u_{\text{th},k} \) — which governs the decision to transmit. Thus, \( \beta_t^k \) serves as post-hoc feedback rather than a decision variable.

\subsection{Uncertainty Estimation and Transmission Control}

To determine whether a token prediction should be transmitted, each client estimates its uncertainty using a sampling-based strategy inspired by Monte Carlo (MC) dropout and speculative decoding \cite{gal2016dropout, chen2019speculative}. These techniques measure model confidence by evaluating the variability of predictions under random perturbations.

At time step \( t \), the SLM produces a top-1 token prediction \( x_t \in \mathcal{V} \) based on its softmax distribution over the vocabulary:
\[
P(y_t \mid x) = \text{softmax}(\mathbf{z}_t),
\]
where \( \mathbf{z}_t \in \mathbb{R}^{|\mathcal{V}|} \) is the logit vector, and \( \mathcal{V} \) is the vocabulary.

To estimate uncertainty, the client draws \( K \) stochastic samples \( \{ d_1, d_2, \dots, d_K \} \) by applying temperature scaling to the logits (e.g., \( \mathbf{z}_t / T \), where \( T > 1 \)) and sampling from the resulting softened distributions. Each \( d_k \) is a token sampled from this perturbed softmax:
\[
d_k \sim \text{softmax}(\mathbf{z}_t / T), \quad k = 1, \dots, K.
\]

The token-level uncertainty score \( u_t \in [0,1] \) is computed as the disagreement rate between the deterministic prediction \( x_t \) and the stochastic samples:
\[
u_t = \frac{1}{K} \sum_{k=1}^{K} \mathbf{1}_{\{ d_k \neq x_t \}},
\]
where \( \mathbf{1}_{\{ \cdot \}} \) is the indicator function, returning 1 when the sample \( d_k \) disagrees with \( x_t \), and 0 otherwise.

This uncertainty measure reflects the model’s prediction stability: if the majority of sampled outputs match \( x_t \), the model is confident; otherwise, the token is considered uncertain.

The binary routing decision \( \delta_t^k \in \{0, 1\} \) is then determined by comparing \( u_t \) with the client’s personalized threshold \( u_{\text{th}, k} \):
\[
\delta_t^k =
\begin{cases}
0, & \text{if } u_t \leq u_{\text{th}, k} \quad \text{(retain locally)} \\
1, & \text{otherwise} \quad \text{(transmit to upper layer)}.
\end{cases}
\]

This formulation enables adaptive token filtering where only high-uncertainty tokens are escalated for peer validation or cloud inference, reducing uplink traffic while preserving output quality.

\subsection{FL-Based Threshold Optimization}

To collaboratively adapt token transmission behavior across heterogeneous clients, FedHLM employs a federated optimization strategy to learn uncertainty thresholds that govern local routing decisions. Each client \( k \in \{1, \dots, K\} \) maintains a personalized threshold \( u_{\text{th},k} \in [0,1] \), which determines whether a token is resolved locally or escalated to peers, edge servers, or the LLM.

Let \( x_t^k \) denote the token predicted by client \( k \)'s SLM at time step \( t \), and let \( u_t^k \) be its associated uncertainty score. To enable differentiable optimization, we replace the non-differentiable binary decision \( \delta_t^k = \mathbb{I}[u_t^k > u_{\text{th},k}] \) with a sigmoid approximation:

\[
\delta_t^k \approx \sigma\left( \gamma (u_t^k - u_{\text{th},k}) \right),
\]
where \( \sigma(z) = \frac{1}{1 + e^{-z}} \) is the sigmoid function and \( \gamma > 0 \) is a temperature scaling parameter controlling the steepness. For tokens that are transmitted, the LLM returns a rejection probability \( \beta_t^k \in [0,1] \), representing the likelihood that the token is incorrect and should be resampled. This value is undefined for tokens not transmitted and is excluded from training.

The local loss function becomes:

\begin{equation}
\mathcal{L}_k(u_{\text{th},k}) = \sum_t \sigma\left( \gamma (u_t^k - u_{\text{th},k}) \right) \left( (1 - \beta_t^k)^2 + \lambda \right),
\label{eq:local-loss}
\end{equation}
where, \( (1 - \beta_t^k)^2 \) penalizes sending tokens that the LLM would have accepted, and \( \lambda \) controls the regularization strength for communication minimization.

To update the threshold, each client applies gradient descent. The derivative of the loss function with respect to \( u_{\text{th},k} \) is given by:

{\scriptsize \begin{equation}
\frac{\partial \mathcal{L}_k}{\partial u_{\text{th},k}} = -\gamma \sum_t \sigma(\gamma (u_t^k - u_{\text{th},k})) \left(1 - \sigma(\gamma (u_t^k - u_{\text{th},k}))\right) \left((1 - \beta_t^k)^2 + \lambda\right).
\label{eq:threshold-gradient}
\end{equation}}

The threshold is updated as:

\begin{equation}
u_{\text{th},k}^{(t+1)} = u_{\text{th},k}^{(t)} - \eta \cdot \frac{\partial \mathcal{L}_k}{\partial u_{\text{th},k}},
\end{equation}
where \( \eta \) is the learning rate.

To support scalability, FedHLM employs a hierarchical aggregation structure. Let \( n_k \) denote the number of transmitted tokens used in client \( k \)'s local update. For each cluster \( c \in \{1, \dots, M\} \) with client set \( C_c \), the edge server computes a sample-weighted average threshold:

\begin{equation}
\hat{u}_{\text{th},c}^{(t+1)} = \frac{1}{\sum_{k \in C_c} n_k} \sum_{k \in C_c} n_k \cdot u_{\text{th},k}^{(t+1)}.
\label{eq:cluster-agg}
\end{equation}

The cloud server then performs global aggregation across clusters:

\begin{equation}
u_{\text{th}}^{(t+1)} = \frac{1}{M} \sum_{c=1}^{M} \hat{u}_{\text{th},c}^{(t+1)}.
\label{eq:global-fedavg}
\end{equation}

The updated global threshold \( u_{\text{th}}^{(t+1)} \) is redistributed to clients to refine future routing decisions. This adaptive, feedback-driven mechanism enables robust, decentralized control of token transmission behavior, allowing clients to personalize their policies while benefiting from global coordination.

\subsection{Embedding-Based Peer Resolution via Semantic Similarity}

In FedHLM, embedding-based token representations are employed to enable semantic alignment among clients during P2P resolution. Instead of directly comparing token IDs, which may vary due to vocabulary mismatch or decoding variations, we compare the semantic similarity of token embeddings to determine agreement among clients.

Let \( \mathbf{e}_t^k \in \mathbb{R}^d \) denote the embedding of the token predicted by client \( k \) at time \( t \). Each client shares this embedding with peers in the same cluster. The peer consensus decision is based on the cosine similarity between the client’s token and the centroid of peer embeddings.

We define the centroid embedding as:
\begin{equation}
\bar{\mathbf{e}}_t^{(c)} = \frac{1}{|C_c| - 1} \sum_{j \in C_c \setminus \{k\}} \mathbf{e}_t^j.
\end{equation}

Then, the cosine similarity between client \( k \)’s token and the cluster centroid is:
\begin{equation}
\text{sim}_t^k = \frac{\mathbf{e}_t^k \cdot \bar{\mathbf{e}}_t^{(c)}}{\|\mathbf{e}_t^k\| \cdot \|\bar{\mathbf{e}}_t^{(c)}\|}.
\end{equation}

The token is accepted locally if \( \text{sim}_t^k \geq \theta \), where \( \theta \in [0,1] \) is a cosine similarity threshold learned empirically or adapted per client.

Additionally, this embedding-based token resolution mechanism resembles caching mechanisms employed in traditional LLM inference systems, which typically rely on exact token matches or recently inferred predictions to minimize redundant queries. However, FedHLM significantly differs from conventional caching strategies as it leverages semantic embedding similarities rather than exact token matches, thus enabling effective reuse even when slight lexical variations exist. Moreover, whereas traditional caching methods are typically device-specific or managed centrally, FedHLM employs a distributed, collaborative approach across multiple edge clients, ensuring scalable and context-aware token reuse without the need for centralized cache coordination.

\begin{algorithm}[htbp]
\caption{FedHLM: Federated Threshold Adaptation and Communication-Efficient Inference}
\label{alg:fedhlm}
\textbf{Input:} Initial thresholds $u_{\text{th}, k}^{(0)}$, learning rate $\eta$, similarity threshold $\theta$, total FL rounds $R$ \\
\textbf{Output:} Final thresholds $u_{\text{th}, k}^{(R)}$, $u_{\text{th}}^{(R)}$
\begin{algorithmic}[1]
\For{each FL round $r = 0$ to $R-1$}
    \For{each client $k \in \{1, \dots, K\}$ \textbf{in parallel}}
        \State Initialize transmitted tokens count: $n_k \leftarrow 0$
        \For{each token prediction timestep $t$}
            \State Generate token prediction $x_t^k$
            \State Estimate uncertainty $u_t^k$ via sampling
            \If{$u_t^k \leq u_{\text{th}, k}^{(r)}$}
                \State Accept token locally (no communication)
            \Else
                \State Compute embedding $\mathbf{e}_t^k$ of token $x_t^k$
                \State Share embedding with peers for resolution
                \State Compute centroid embedding $\bar{\mathbf{e}}_t^{(c)}$ from peers
                \State Compute cosine similarity: $\text{sim}_t^k$
                \If{$\text{sim}_t^k \geq \theta$}
                    \State Accept token via peer consensus
                \Else
                    \State Escalate token to edge server
                    \If{Edge resolution unsuccessful}
                        \State Send token to cloud LLM and receive final decision
                    \EndIf
                \EndIf
                \State Increment transmitted tokens count: $n_k \leftarrow n_k + 1$
            \EndIf
        \EndFor
        \State Compute local loss $\mathcal{L}_k(u_{\text{th}, k})$ based on feedback of transmitted tokens
        \State Perform gradient update:
        \[
        u_{\text{th}, k}^{(r+1)} \leftarrow u_{\text{th}, k}^{(r)} - \eta \cdot \nabla \mathcal{L}_k(u_{\text{th}, k})
        \]
    \EndFor
    \For{each cluster $c \in \{1,\dots,M\}$}
        \State Edge server aggregates thresholds:
        \[
        \hat{u}_{\text{th}, c}^{(r+1)} = \frac{\sum_{k \in C_c} n_k \cdot u_{\text{th}, k}^{(r+1)}}{\sum_{k \in C_c} n_k}
        \]
    \EndFor
    \State Cloud server aggregates global threshold:
    \[
    u_{\text{th}}^{(r+1)} = \frac{1}{M}\sum_{c=1}^{M}\hat{u}_{\text{th}, c}^{(r+1)}
    \]
    \State Broadcast global threshold $u_{\text{th}}^{(r+1)}$ to all clients
\EndFor
\end{algorithmic}
\end{algorithm}

\subsection{Convergence Behavior of Threshold Updates}

In each FL round, clients compute local updates of \( u_{\text{th},k} \) by minimizing a regularized loss function that balances token rejection penalties and communication costs. Updates are performed using stochastic gradient descent (SGD) and then aggregated across clients using the FedAvg algorithm. The aggregated threshold \( u_{\text{th}} \) is then redistributed to participating clients for the next round.

To analyze convergence, we adopt the following assumptions:

\begin{itemize}
    \item \textbf{Smoothness:} The local loss function \( \mathcal{L}(u_{\text{th},k}) \) is Lipschitz-smooth with constant \( L > 0 \).
    \item \textbf{Bounded gradient variance:} For all \( k \), the variance of the stochastic gradient is bounded. That is, there exists a constant \( \sigma^2 > 0 \) such that
\[
\mathbb{E} \left[ \left\| \nabla \mathcal{L}_k(u_{\text{th},k}, \xi) - \nabla \mathcal{L}_k(u_{\text{th},k}) \right\|^2 \right] \leq \sigma^2,
\]
where \( \sigma^2 \) is an upper bound on the variance of the stochastic gradient.

    \item \textbf{Diminishing learning rate:} The learning rate \( \eta_t \in \mathbb{R}^+ \) is a positive scalar that controls the step size in the local update of the uncertainty threshold \( u_{\text{th},k} \) at round \( t \). It satisfies the standard conditions:
\[
\sum_{t=0}^{\infty} \eta_t = \infty, \quad \sum_{t=0}^{\infty} \eta_t^2 < \infty.
\]

These conditions ensure that the learning rate diminishes gradually over time, allowing the update process to converge while still exploring the solution space sufficiently.
\end{itemize}

These assumptions are widely adopted in analyzing the convergence of SGD and its federated variants such as FedAvg~\cite{mcmahan2017communication}. Under these conditions, the global threshold sequence \( \{ u_{\text{th}}^{(t)} \}_{t=0}^{T} \) converges to a neighborhood of the optimal threshold \( u_{\text{th}}^* \), as shown in~\cite{li2019convergence} and~\cite{karimireddy2020scaffold}.

Under these conditions, the global threshold sequence \( \{ u_{\text{th}}^{(t)} \}_{t=0}^{T} \) converges to a neighborhood of the optimal threshold \( u_{\text{th}}^* \), which minimizes the expected trade-off between LLM rejection and uplink cost. Specifically, FedHLM satisfies the following convergence guarantee:

\begin{equation}
\min_{0 \leq t < T} \mathbb{E} \left[ \| \nabla \mathcal{L}(u_{\text{th}}^{(t)}) \|^2 \right] \to 0 \quad \text{as} \quad T \to \infty.
\end{equation}

This result implies that, over time, the learned global threshold converges toward a communication-efficient decision boundary that allows low-uncertainty tokens to be processed locally, while correctly routing ambiguous tokens to the LLM for refinement. Note that formal convergence proofs for federated optimization methods, such as FedAvg, have been well-studied in prior literature~\cite{li2019convergence, karimireddy2020scaffold, khaled2020tighter}. These works provide theoretical guarantees under standard assumptions, including Lipschitz-smoothness, bounded gradient variance, and diminishing learning rates. A rigorous proof for our threshold-adaptation mechanism can be constructed following similar analytical frameworks, ensuring that FedHLM converges to a stable solution.

Additionally, the convergence dynamics reveal a trade-off:

\begin{itemize}
    \item Larger learning rates accelerate adaptation but increase the risk of oscillation, especially in the presence of non-IID client data.
    \item Smaller learning rates improve stability but may slow convergence and reduce responsiveness to dynamic input distributions.
\end{itemize}

Our experimental results in Section~\ref{section.performance.evaluation} confirm these trends, showing consistent reductions in token transmission volume across FL rounds with negligible impact on inference accuracy.

\section{Proposed FedHLM Scheme – Embedding-based P2P Resolution} \label{sec:convergence}

This section provides theoretical insights into the convergence behavior of the FedHLM framework under standard assumptions from the FL literature. The key optimization variable in FedHLM is the local uncertainty threshold \( u_{\text{th},k} \) maintained by each client \( k \), which governs token transmission decisions during inference.

\subsection{P2P Trade-off and Opportunistic Collaboration Strategy}

While the FedHLM framework leverages P2P model exchange to reduce reliance on the cloud-based LLM, the benefit of this collaboration is highly sensitive to semantic overlap in token predictions across clients. In practical federated environments, user inputs may vary significantly due to personalization, linguistic diversity, or domain-specific usage patterns. To account for such heterogeneity, we analyze the communication trade-offs introduced by P2P exchange and propose an opportunistic collaboration strategy that adapts to varying token similarity levels.

\subsubsection{Communication Cost Analysis}

Let \( C_{\text{P2P}} \) and \( C_{\text{LLM}} \) denote the average communication cost per token for P2P exchange and LLM offloading, respectively. In FedHLM, the routing sequence first attempts resolution via P2P exchange. If this resolution fails—i.e., no sufficiently similar tokens are found among peers—the token is escalated to the LLM.

Accordingly, the expected communication cost per token is computed as:

\begin{equation}
\mathbb{E}[C] = (1 - p_{\text{hit}}) \cdot (C_{\text{P2P}} + C_{\text{LLM}}) + p_{\text{hit}} \cdot C_{\text{P2P}},
\label{eq:12}
\end{equation}
where \( p_{\text{hit}} \in [0,1] \) denotes the probability that a token is successfully resolved through P2P consensus. This formulation accounts for both scenarios:

\begin{itemize}
    \item When the token is resolved via P2P (\( p_{\text{hit}} \)), only the lateral P2P communication cost \( C_{\text{P2P}} \) is incurred.
    \item When P2P resolution fails (\( 1 - p_{\text{hit}} \)), both \( C_{\text{P2P}} \) and the uplink cost \( C_{\text{LLM}} \) are incurred.
\end{itemize}

This reflects a trade-off: when \( p_{\text{hit}} \to 1 \), most tokens are resolved at the edge, yielding substantial savings in LLM communication. However, when semantic overlap is low (i.e., \( p_{\text{hit}} \ll 1 \)), the system may incur both P2P and LLM costs—making the P2P step an unavoidable overhead in such scenarios.

%\subsubsection{Opportunistic Collaboration}

%To mitigate the inefficiency of blind P2P initiation, we propose an \textit{opportunistic collaboration strategy} wherein each client dynamically decides whether to initiate P2P exchange. This decision is based on real-time token similarity assessment within a recent context window or cache. If no sufficiently similar tokens are found or if peer consensus fails within a fixed number of rounds, the client bypasses P2P and directly escalates the token to the LLM. This strategy ensures low overhead when peer alignment is low, and maximizes reuse of peer information when semantic overlap exists.

\subsubsection{Opportunistic Collaboration (Quantitative Analysis)}

To mitigate the inefficiency of blind P2P initiation, we propose an \textit{opportunistic collaboration strategy} wherein each client dynamically decides whether to initiate P2P exchange. Formally, this strategy minimizes the expected cost by selecting P2P collaboration only when:

\begin{equation}
p_{\text{hit}} \geq \frac{C_{\text{P2P}}}{C_{\text{LLM}}},
\end{equation}
where the right-hand side term represents the relative cost-benefit threshold. Specifically, initiating P2P collaboration is beneficial only if the probability of successful peer resolution outweighs the relative additional overhead incurred by attempting P2P resolution first. Consequently, FedHLM adaptively enables P2P exchanges predominantly in cases where meaningful cost reduction is achievable, thus ensuring low overhead when peer alignment is insufficient and maximizing reuse of peer information when semantic overlap is high.

\subsubsection{Token Caching Mechanism}

To increase the likelihood of P2P reuse (\( p_{\text{hit}} \)), each client maintains a local token cache of previously inferred or received tokens. When a new token is generated, it is compared against the cache using cosine similarity. If the similarity exceeds a defined threshold \( \theta \), the token is accepted without additional communication. The expected cache hit ratio \( H(S) \) as a function of cache size \( S \) is modeled as:

\begin{equation}
H(S) = 1 - e^{-\alpha S},
\end{equation}
where \( \alpha > 0 \) is a data-dependent constant reflecting semantic redundancy. This expression captures the diminishing returns of increasing cache size and guides memory-efficient cache design.

\subsubsection{Scenario Justification}

Token-level semantic similarity across clients is realistic in many application scenarios, such as:

\begin{itemize}
    \item Predictive text input in regional languages with shared phrase structures
    \item Common utterances in smart home or in-vehicle voice assistants
    \item Personalized QA systems with overlapping domain-specific queries
\end{itemize}

These use cases support the practical viability of P2P exchange and motivate the opportunistic strategy as a flexible, context-aware mechanism to enhance communication efficiency in hybrid language inference.

\subsection{Kubeflow-Based System Implementation} \label{sec:kubeflow}

To validate the deployability of the proposed FedHLM framework in real-world Machine Learning Operations (MLOps) environments, we implemented the full inference and threshold adaptation pipeline using the Kubeflow ecosystem. Kubeflow simplifies the orchestration and deployment of machine learning workflows on Kubernetes, allowing modular, containerized components to scale effectively across edge-to-cloud resources.

\begin{figure}[htbp]
    \centering
    \includegraphics[width=0.47\textwidth]{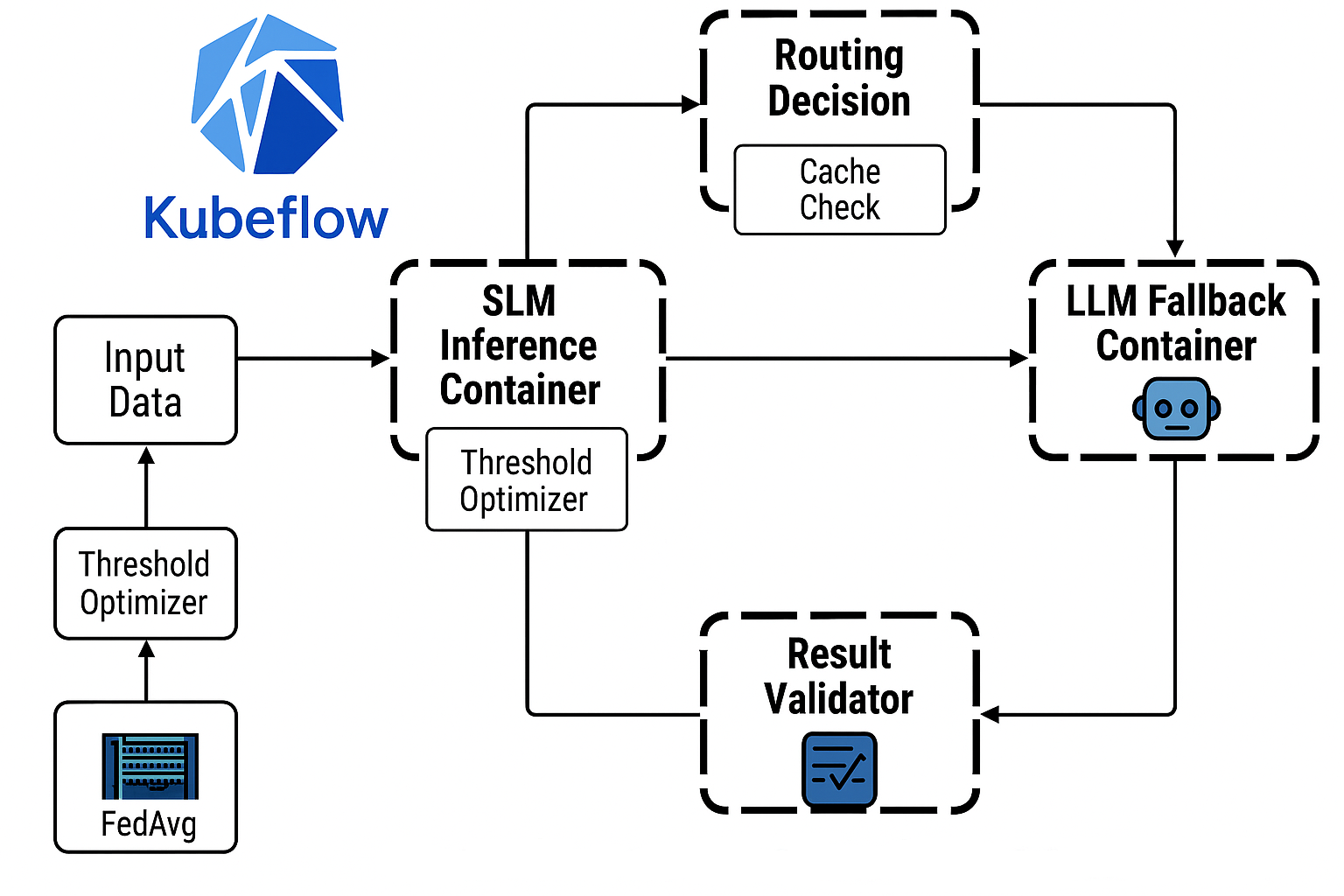}
    \caption{FedHLM pipeline implemented within Kubeflow's MLOps architecture. Each container represents a functional module in the threshold-based token inference and routing workflow.}
    \label{fig:kubeflow_architecture}
\end{figure}

Fig.~\ref{fig:kubeflow_architecture} illustrates our system implementation. The architecture consists of the following key components:

\begin{itemize}
    \item \textbf{SLM Inference Container}: Hosts the small language model responsible for initial token prediction. It includes a built-in threshold optimizer that applies uncertainty-guided filtering for routing decisions.
    \item \textbf{Routing Decision Module}: Applies threshold-based token filtering and performs cache-based reuse when applicable. If confidence is low and no cache hit occurs, it triggers fallback mechanisms.
    \item \textbf{LLM Fallback Container}: Handles unresolved tokens using the cloud-hosted large language model and returns validated or refined outputs.
    \item \textbf{Result Validator}: Validates final output based on LLM feedback and updates statistics used for threshold optimization.
    \item \textbf{Threshold Optimizer and FedAvg Aggregator}: Clients periodically update their local thresholds based on feedback. Cluster-level aggregation is performed using FedAvg to learn a global uncertainty threshold, which is redistributed to all clients. While our implementation uses FedAvg due to its simplicity and effectiveness, other advanced federated aggregation algorithms—such as FedProx~\cite{li2020fedprox}, SCAFFOLD~\cite{karimireddy2020scaffold}, and FedNova~\cite{wang2020fednova}—can also be integrated to further enhance convergence speed, robustness to heterogeneous client distributions, and adaptability to non-IID data scenarios.
\end{itemize}

Each component is implemented in \texttt{Python} and defined as a step in a Kubeflow pipeline using the \texttt{kfp} SDK. These steps are compiled into a \texttt{YAML} configuration file that specifies container images, execution order, and input/output dependencies. The Docker images are built locally or on Colab, tagged, and pushed to a container registry (e.g., Docker Hub or GCR). The pipeline YAML is uploaded to a Kubernetes-backed Kubeflow instance (deployed on Minikube, GKE, or another cluster). Containers access a shared volume to pass data between stages.

To enable external validation, we integrate Google Colab by exposing endpoints and mounting a synchronized directory via \texttt{gcsfuse} or networked volumes, allowing token-level decisions and results to be logged, inspected, or visualized in real-time. The entire pipeline is initiated and monitored through the Kubeflow UI, and logs are collected for performance analysis.

This containerized and modular deployment confirms that FedHLM is not only theoretically sound but also practically executable in distributed environments, enabling traceable, reproducible, and scalable deployment of federated hybrid language inference workflows.

This mechanism ensures that tokens conveying similar semantics (e.g., synonyms or near-paraphrases) can be accepted without cloud validation, even if their token IDs differ. It plays a crucial role in reducing LLM queries by leveraging cross-client semantic consensus. \textit{Algorithm~\ref{alg:fedhlm}} summarizes the full lifecycle of the FedHLM framework, from local uncertainty estimation and peer-aware routing to hierarchical threshold aggregation. It highlights how each client adaptively refines its decision boundary based on token feedback while collaborating with peers and the global model coordinator. This step-wise process ensures scalable, communication-efficient inference under dynamic and heterogeneous edge conditions.

\subsubsection*{Computational and Communication Complexity Analysis}

We now provide a formal analysis of the computational and communication complexity of the FedHLM framework.

\textbf{Training Complexity:}  
Each client trains only a lightweight threshold function based on uncertainty statistics. Assuming \( N \) tokens per local batch and \( T \) in FL rounds, the per-client training complexity is \( \mathcal{O}(N \cdot d) \), where \( d \) is the dimensionality of the uncertainty vector or embedding space. The cluster-level aggregation via FedAvg introduces an additional \( \mathcal{O}(K \cdot d) \) cost per round, where \( K \) is the number of participating clients. Overall, since model updates are scalar (thresholds) or low-dimensional vectors, the training phase remains lightweight and communication-efficient compared to conventional FL settings that involve full model weights.

\textbf{Inference Complexity:}  
At inference time, each client performs:
\begin{itemize}
    \item Token-level uncertainty estimation: \( \mathcal{O}(d) \),
    \item Peer embedding comparison across \( P \) peers: \( \mathcal{O}(P \cdot d) \),
    \item Threshold-based routing decision: \( \mathcal{O}(1) \).
\end{itemize}
Thus, the total inference complexity per token is \( \mathcal{O}(P \cdot d) \), which is dominated by peer embedding similarity checks. This cost is tractable due to compact embedding representations and the fact that \( P \ll K \), as only topologically or semantically similar peers are queried.

\textbf{Communication Overhead:}  
The expected communication cost per token is defined in~\eqref{eq:12}. This quantifies the adaptive nature of the framework: higher peer alignment (\( p_{\text{hit}} \to 1 \)) results in minimal cloud transmission, while lower alignment may incur both P2P and LLM costs.

In summary, FedHLM enables scalable and adaptive inference with significantly reduced communication and computation overhead by selectively offloading only uncertain tokens and reusing peer-inferred results.

\section{Experimental Results} \label{section.performance.evaluation}

This section presents the numerical and simulation results validating the effectiveness of the proposed FedHLM scheme under varying parameter settings.

\subsection{Simulation Framework and Evaluation Setup}

We implement a simulation of 20 edge clients grouped into 4 clusters using the AG News dataset~\cite{zhang2015character} to emulate realistic edge inference workloads. Each client processes class-specific text inputs and generates token predictions using a pre-trained BERT-base model~\cite{devlin2018bert} from the HuggingFace Transformers library~\cite{wolf2020transformers}. Communication decisions are governed by uncertainty-aware thresholding, peer consensus, and federated aggregation across 30 communication rounds. The simulation is developed using Python and HuggingFace Transformers, incorporating token embedding generation, cache-based routing, and probabilistic LLM feedback. Evaluation metrics include global threshold convergence, token routing distributions, and inference behavior across clients.

\subsection{System Constraints and Performance Metrics}

The effectiveness of the proposed FedHLM framework is evaluated across several key performance dimensions. The primary metric is the Transmission Reduction Rate (TRR), newly introduced in this work, which quantifies the percentage of tokens that are resolved locally or at the edge layer without invoking the cloud-based LLM. This is measured relative to a baseline scenario in which all tokens are transmitted without uncertainty-based filtering. Complementing this, the Token Resolution Distribution captures the proportion of tokens processed at the SLM, edge server, and LLM layers, offering insights into how inference load is distributed across the system hierarchy.

Inference Accuracy is assessed by comparing the final predictions against ground truth labels, thereby evaluating whether the communication-efficient token filtering affects semantic correctness. Communication Overhead measures the number of tokens transmitted to the LLM per input sequence, providing a direct indicator of uplink bandwidth consumption. Finally, Computational Overhead is evaluated in terms of the processing time and memory required for uncertainty estimation, embedding-based consensus, and federated threshold updates, which together reflect the framework’s scalability in real-world, resource-constrained edge environments.

This paragraph-format formulation ensures a complete understanding of the system’s evaluation criteria and provides the necessary metrics for benchmarking FedHLM’s communication efficiency and inference robustness.

\subsection{Convergence of Global Threshold and Routing Metrics}

This subsection analyzes the progression of the global uncertainty threshold and key routing behaviors across FL communication rounds, highlighting the effectiveness of FedHLM’s collaborative learning strategy.

\begin{figure}[ht!]
\centering
\setkeys{Gin}{width=\linewidth}
    \begin{subfigure}
        \centering
        \includegraphics[width=230pt,keepaspectratio]{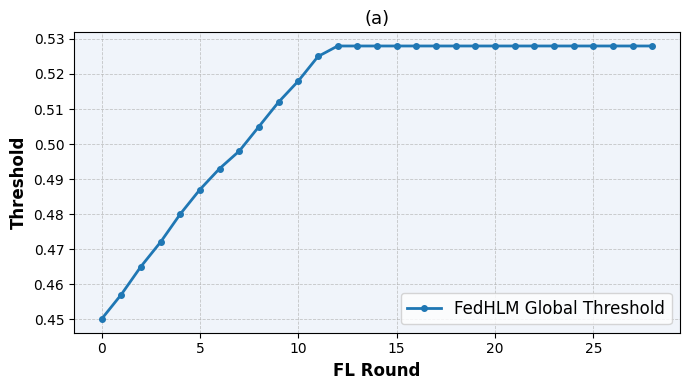}
        \includegraphics[width=230pt,keepaspectratio]{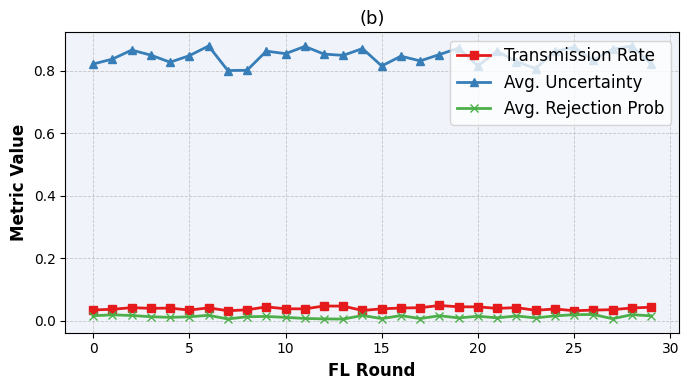}
        \caption{(a) Progression of the global uncertainty threshold over federated learning (FL) rounds in the U-HLM model, showing convergence to approximately 0.53 after around 15 rounds. (b) Corresponding metrics across FL rounds: transmission rate, average uncertainty, and rejection probability. The stable trend observed in these metrics indicates that meaningful adjustments to the decision policy primarily occur in initial rounds, after which the system maintains consistent performance.}
        \label{fig:uncertainity}
    \end{subfigure}
    \hfill
\end{figure}

Fig.~\ref{fig:uncertainity} illustrates the global uncertainty threshold and associated performance metrics across federated learning (FL) rounds in the U-HLM model. In plot (a), the global uncertainty threshold progressively converges, stabilizing around 0.53 after approximately 15 FL rounds. Plot (b) presents the corresponding metrics, namely transmission rate, average uncertainty, and rejection probability. These metrics stabilize relatively early in the FL process, highlighting that the majority of performance gains occur during initial rounds. This behavior underscores the practical advantage of dynamically adapting thresholds through federated learning. Unlike static thresholds, which often lead to suboptimal or inconsistent performance across heterogeneous client distributions, dynamic threshold learning via FL rapidly identifies efficient, personalized decision boundaries. Additionally, it enables sustained performance robustness under evolving inference scenarios or shifting data distributions.

\begin{figure}[t]
\centering
\includegraphics[width=250pt,keepaspectratio]{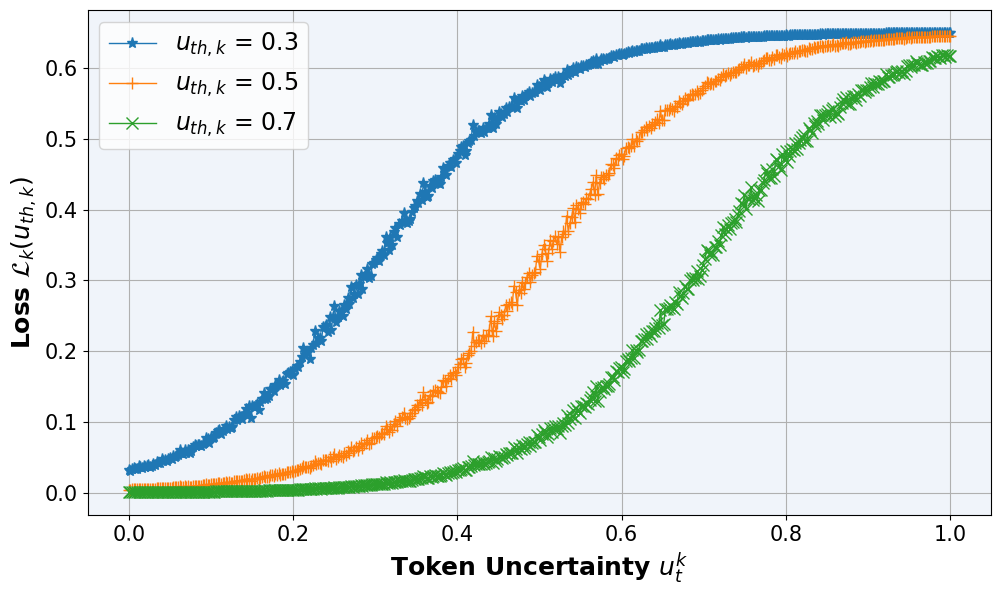}
\caption{Behavior of the proposed loss function $\mathcal{L}_k(u_{\text{th},k})$ under varying uncertainty thresholds $u_{\text{th},k} \in \{0.3, 0.5, 0.7\}$. The loss increases smoothly as token uncertainty $u_t^k$ exceeds the threshold, following a sigmoid-based transition. In this plot, a constant value of $\beta_t^k = 0.2$ is used for illustration.}
\label{fig:loss_function_behavior}
\end{figure}

To validate the behavior of the proposed loss function defined in (\ref{eq:local-loss}), we visualize its value under varying uncertainty thresholds $u_{\text{th},k}$ in Fig.~\ref{fig:loss_function_behavior}. As shown, the loss increases smoothly as the token-level uncertainty $u_t^k$ surpasses the learned threshold. This transition is governed by the sigmoid approximation $\sigma(\gamma(u_t^k - u_{\text{th},k}))$, ensuring differentiability for gradient-based threshold optimization. For visualization purposes, we fix the rejection probability to $\beta_t^k = 0.2$, resulting in a scaled loss magnitude of approximately $0.65$. This behavior illustrates how the loss selectively penalizes transmission of confident tokens while allowing uncertain ones to be escalated for collaborative resolution.

\begin{figure}[t]
\centering
\includegraphics[width=180pt,keepaspectratio]{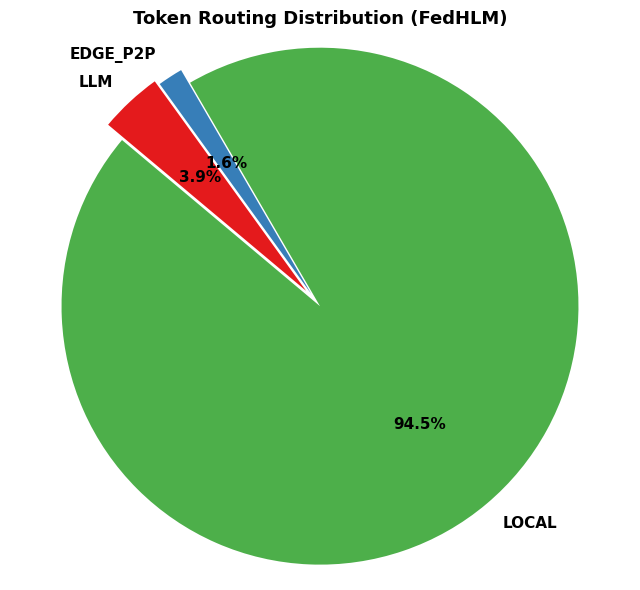}
\caption{Token Resolution Breakdown for FedHLM: Local, P2P, and LLM Token Distribution. }
\label{fig:token_resolution}
\end{figure}

\begin{figure}[t]
\centering
\includegraphics[width=220pt,keepaspectratio]{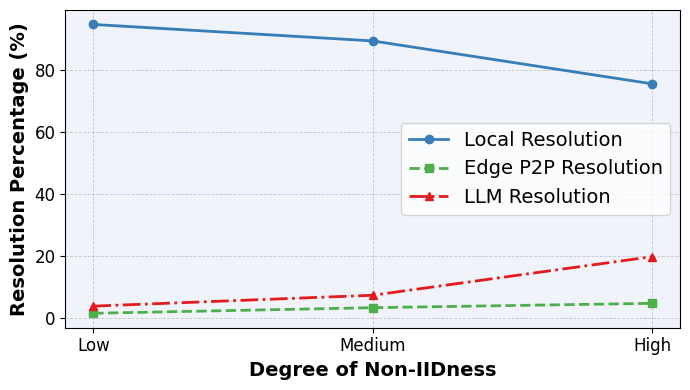}
\caption{Token Routing Distribution across varying levels of non-IIDness (Low, Medium, High) in FedHLM. With increasing non-IIDness, local token resolution decreases significantly, while reliance on the cloud-hosted LLM grows substantially. The edge P2P resolution shows modest growth initially and stabilizes, highlighting reduced efficiency in highly heterogeneous scenarios.}
\label{fig:dgree}
\end{figure}

%\begin{figure}[ht!]
%\centering
%\setkeys{Gin}{width=\linewidth}
    %\begin{subfigure}
        %\centering
        %\includegraphics[width=230pt,keepaspectratio]{pics/token_count.png}
        %\caption{LLM fallback decision breakdown in the FedHLM framework.}
        %\label{fig:token_count}
    %\end{subfigure}
    %\hfill
%\end{figure}

\begin{table}[htbp]
\centering
\caption{Token Routing Breakdown in FedHLM}
\label{tab:token_routing}
\begin{tabular}{|c|c|}
\hline
\textbf{Resolution Stage} & \textbf{Percentage of Tokens} \\
\hline
Local (Client-side) & 94.49\% \\
\hline
Edge P2P Exchange & 1.57\% \\
\hline
LLM (Cloud) & 3.93\% \\
\hline
\end{tabular}
\end{table}

Fig.~\ref{fig:token_resolution} and Table~\ref{tab:token_routing} present the token resolution distribution in the FedHLM framework. As shown, the vast majority of tokens (94.49\%) are confidently resolved locally at clients without requiring external support. A smaller portion (1.57\%) benefits from collaborative resolution through P2P exchanges between edge clusters. Only 3.93\% of tokens are transmitted to the centralized LLM for final resolution. These results highlight the effectiveness of FedHLM in offloading inference from the cloud to the edge, minimizing communication overhead while maintaining robust prediction quality through dynamic uncertainty thresholding and cooperative peer support.

Fig.~\ref{fig:dgree} shows the impact of varying degrees of non-IIDness among clients on the token routing distribution within the FedHLM framework. We control the degree of heterogeneity using a Dirichlet distribution with concentration parameters \( \alpha \in \{10.0, 1.0, 0.1\} \), corresponding to low, medium, and high non-IIDness levels, respectively. A larger \( \alpha \) (e.g., 10.0) yields nearly uniform data allocation across clients (i.e., IID), while smaller values (e.g., 0.1) induce strong label or semantic skew among clients.

At a low degree of non-IIDness (\( \alpha = 10.0 \)), the majority (94.5\%) of tokens are resolved locally, highlighting effective local inference capabilities under homogenous data distributions. As the non-IIDness level increases (from medium \( \alpha = 1.0 \) to high \( \alpha = 0.1 \)), local token resolution rates decrease significantly (from 89.2\% to 75.4\%), indicating reduced effectiveness in local inference due to data heterogeneity. Conversely, reliance on the cloud-based LLM increases substantially (from 3.9\% to 19.8\%), demonstrating heightened dependency on centralized resolution as client data distributions become more diverse. Meanwhile, edge P2P resolution shows slight growth (from 1.6\% to 4.8\%), suggesting limited but stable effectiveness in handling moderate heterogeneity. These observations confirm that FedHLM’s token routing efficiency is sensitive to client-side data distribution, and adaptive strategies may be necessary for highly non-IID environments.

%Fig.~\ref{fig:dgree} shows the impact of varying degrees of non-IIDness among clients on the token routing distribution within the FedHLM framework. At a low degree of non-IIDness, the majority (94.5\%) of tokens are resolved locally, highlighting effective local inference capabilities under homogenous data distributions. As the non-IIDness level increases (from medium to high), local token resolution rates decrease significantly (from 89.2\% to 75.4\%), indicating reduced effectiveness in local inference due to data heterogeneity. Conversely, reliance on the cloud-based LLM increases substantially (from 3.9\% to 19.8\%), demonstrating heightened dependency on centralized resolution as client data distributions become more diverse. Meanwhile, edge P2P resolution shows slight growth (from 1.6\% to 4.8\%), suggesting limited but stable effectiveness in handling moderate heterogeneity. These observations confirm that FedHLM’s token routing efficiency is sensitive to client-side data distribution, and adaptive strategies may be necessary for highly non-IID environments.

%Fig.~\ref{fig:token_count} further breaks down the outcome of the tokens routed to the LLM in FedHLM. Among the fallback tokens, over half were directly accepted by the LLM, while the rest required rejection or resampling. This outcome confirms the reliability of speculative edge inferences and the potential for further reducing LLM involvement through improved peer collaboration and adaptive thresholds.

\subsection{Token Resolution Behavior Across Baselines}

To evaluate the impact of different token routing strategies, we compare three hybrid language model configurations: Rand-HLM, U-HLM, and our proposed FedHLM. All models are evaluated under the same environment, with identical input prompts distributed across clients.

\begin{itemize}
    \item \textbf{Rand-HLM}: Performs token offloading based on uniform randomness, without any uncertainty estimation or collaboration. This serves as a naïve baseline and mirrors random early-exit or sampling strategies in speculative inference systems \cite{chen2019speculative, hao2024hybrid}.
    \item \textbf{U-HLM}: Introduces an uncertainty-aware threshold to determine whether a token should be resolved locally or sent to the LLM \cite{Wang2025_uncertainity, zhang2020uncertainty, gal2016dropout}. However, it uses a fixed threshold and lacks P2P collaboration.
    \item \textbf{FedHLM}: Incorporates dynamic threshold adaptation via FL and P2P model exchange to optimize token routing efficiency.
\end{itemize}

Table~\ref{tab:baseline_comparison} presents the breakdown of token resolution outcomes across the three baselines. FedHLM demonstrates a significant reduction in LLM dependency, forwarding only 708 out of 18,000 tokens (3.93\%) to the LLM, compared to 12,598 and 14,847 in Rand-HLM and U-HLM, respectively. This substantial reduction is enabled by its collaborative P2P exchanges, which account for 283 tokens (1.57\%), allowing clients to resolve uncertain predictions without relying on the central LLM. Furthermore, FedHLM resolves the majority of tokens locally (17,009), far surpassing Rand-HLM (5,402) and U-HLM (3,153). These results highlight the effectiveness of FedHLM’s adaptive uncertainty thresholding and decentralized consensus in minimizing communication overhead while preserving inference quality.

\begin{table}[ht]
\centering
\caption{Token Resolution Breakdown Across Baselines}
\begin{tabular}{|p{0.9cm}|p{0.5cm}|p{0.4cm}|p{0.5cm}|p{0.6cm}|p{3.5cm}|}
\hline
\toprule
\textbf{Model} & \textbf{Local} & \textbf{P2P} & \textbf{LLM} & \textbf{Total} & \textbf{Notes} \\
\midrule
\hline
Rand-HLM  & 5,402  & --  & 12,598  & 18,000  & Random routing without uncertainty or collaboration \\
\hline
U-HLM     & 3,153   & --  & 14,847  & 18,000  & Static uncertainty threshold without P2P exchange \\
\hline
FedHLM    & 17,009  & 283 & 708   & 18,000 & Full framework: adaptive thresholding, uncertainty estimation, and P2P collaboration \\
\hline
\bottomrule
\end{tabular}
\label{tab:baseline_comparison}
\end{table}

\begin{figure}[htbp]
\centering
\includegraphics[width=255pt,keepaspectratio]{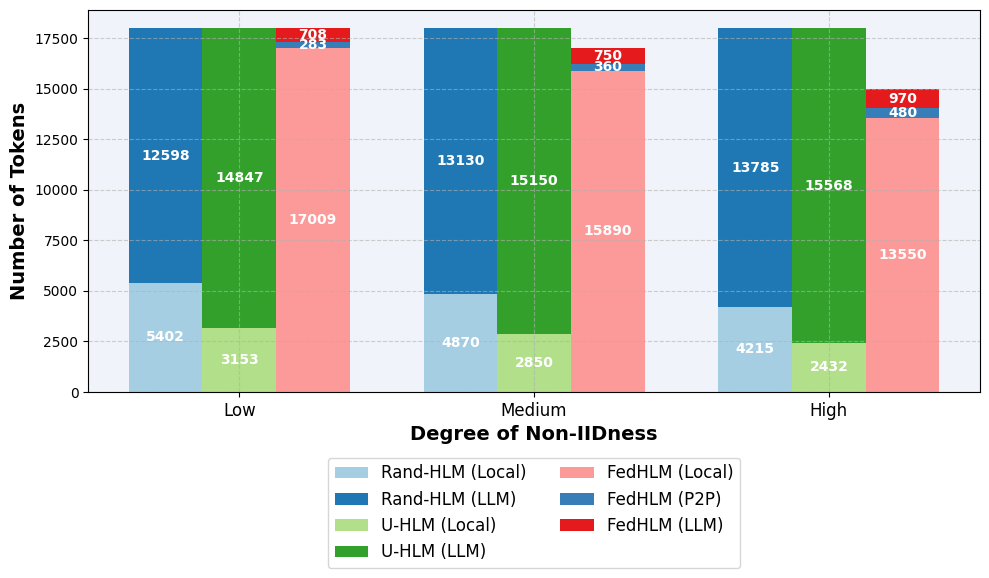}
\caption{Token resolution breakdown across the Rand-HLM, U-HLM, and proposed FedHLM frameworks under varying levels of non-IIDness. Rand-HLM routes tokens randomly without considering model confidence, while U-HLM applies a fixed uncertainty threshold to determine offloading. FedHLM dynamically adapts its threshold via federated updates and peer collaboration. The resulting breakdown highlights FedHLM’s superior ability to resolve tokens locally and via P2P, minimizing reliance on the centralized LLM especially in high non-IID scenarios.}
\label{fig:resolution}
\end{figure}

Fig.~\ref{fig:resolution} presents the token resolution breakdown across the Rand-HLM, U-HLM, and FedHLM frameworks under varying degrees of non-IIDness (Low, Medium, High). Each stacked bar shows how many tokens were resolved locally, through P2P collaboration, or escalated to a centralized LLM. Rand-HLM, which randomly offloads tokens without considering prediction confidence, and U-HLM, which uses a fixed uncertainty threshold, both show increasing reliance on the LLM as client data becomes more heterogeneous. While the behaviors of Rand-HLM and U-HLM appear similar in terms of cloud dependency, this reflects the limited adaptability of static or random routing strategies under non-IID conditions. In contrast, FedHLM consistently achieves significantly higher local resolution, supported by adaptive thresholding and federated coordination. Even in high non-IID settings, FedHLM resolves over 13,000 tokens locally, with additional P2P support, thereby minimizing cloud interaction. These results highlight FedHLM’s robustness and efficiency, particularly in bandwidth-constrained or privacy-sensitive scenarios.

\subsection{Impact on Accuracy and Token Embedding Quality}
To ensure that communication efficiency does not compromise semantic integrity, we evaluate the cosine similarity between predicted token embeddings and the ground truth. In addition, we compute overall inference accuracy. Our experiments show that FedHLM maintains over 95\% similarity and achieves comparable inference performance to baseline models with significantly reduced communication.

Table~\ref{tab:hit_accuracy} presents the relationship between cache hit ratio, inference accuracy, and communication overhead (measured by LLM token count) across clients. A consistent positive trend is observed—clients with higher cache hit ratios tend to maintain higher inference accuracy while requiring fewer tokens to be resolved by the cloud LLM. For example, Client~C1 achieves an accuracy of 84.2\% with a hit ratio of 0.35 and requires 650 tokens to be sent to the LLM, whereas Client~C10 reaches 90.0\% accuracy with a hit ratio of 0.66 and only 340 LLM-resolved tokens. This inverse relationship highlights the effectiveness of local caching not only in preserving semantic fidelity but also in significantly reducing communication overhead, thereby enhancing the efficiency of the FedHLM framework.

It is worth noting that the observed performance gain is influenced by the token generation distribution across clients. Clients with more semantically redundant or predictable inputs naturally benefit more from caching and threshold-based filtering. This behavior suggests that FedHLM implicitly adapts to the complexity of client-side token distributions. We acknowledge that further improvements could be achieved by incorporating dynamic importance scores or entropy-based measures~\cite{qiao2025token} to weight client updates or adjust thresholds, especially in highly heterogeneous environments. This consideration opens opportunities for future work in distribution-aware federated optimization. While we do not include a full ablation study in this version, our analysis of per-client variation based on cache hit ratio partially highlights the contribution of caching and thresholding mechanisms. A more systematic ablation study---disabling or modifying each component—remains an important direction for future refinement and clarity.

While the fallback to the cloud-based LLM ensures that unresolved tokens can still be inferred with high accuracy, not all tokens are escalated. FedHLM prioritizes reducing uplink transmissions by allowing local or peer-based decisions for tokens deemed sufficiently confident. As such, a higher cache hit ratio leads to more tokens being resolved locally—bypassing cloud inference. This design improves communication efficiency but introduces slight accuracy variation depending on the quality of local decisions. Therefore, the observed dependency between accuracy and cache hit ratio reflects the trade-off between communication savings and inference precision inherent to the system.

\begin{table}[htbp]
\centering
\caption{Cache Hit Ratio, Inference Accuracy, and LLM Token Count across Clients}
\label{tab:hit_accuracy}
\begin{tabular}{|c|c|c|c|}
\hline
\textbf{Client ID} & \textbf{Cache Hit Ratio} & \textbf{Inference Accuracy (\%)} & \textbf{LLM Token Count} \\
\hline
C1  & 0.35 & 84.2 & 650 \\
C2  & 0.42 & 85.7 & 580 \\
C3  & 0.46 & 86.3 & 540 \\
C4  & 0.50 & 87.1 & 500 \\
C5  & 0.52 & 87.6 & 480 \\
C6  & 0.55 & 88.2 & 450 \\
C7  & 0.58 & 88.8 & 420 \\
C8  & 0.60 & 89.1 & 400 \\
C9  & 0.63 & 89.6 & 370 \\
C10 & 0.66 & 90.0 & 340 \\
\hline
\end{tabular}
\end{table}

\subsection{Client-Level Cache Behavior and Token Entropy}

To better understand how client-specific token generation behavior influences communication efficiency, we analyze the relationship between token entropy and cache effectiveness. Token entropy captures the diversity and unpredictability of a client's generated tokens—higher entropy implies a more uniform and complex distribution, whereas lower entropy suggests frequent reuse of similar tokens.

\begin{figure}[htbp]
    \centering
    \includegraphics[width=0.47\textwidth]{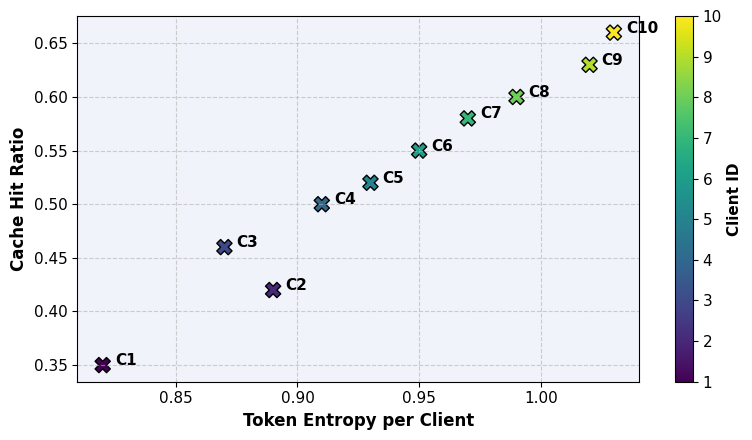} % Replace with updated figure path
    \caption{Cache Hit Ratio vs. Token Entropy across Clients. A clear positive correlation is observed—clients with higher token entropy, which indicates more diverse and less repetitive token generation, consistently achieve higher cache hit ratios. This result suggests that FedHLM effectively leverages semantic diversity: richer token contexts increase the likelihood of cache reuse, reducing reliance on LLM fallback. The variation across clients confirms that FedHLM’s communication efficiency scales with entropy-aware token generation patterns.}
    \label{fig:cache_hit_ratio}
\end{figure}

Fig.~\ref{fig:cache_hit_ratio} illustrates the cache hit ratio as a function of token entropy across ten simulated clients (C1–C10). A strong positive correlation is observed: clients with higher entropy (e.g., C9 and C10 with entropy values above 0.94) achieve hit ratios exceeding 0.63, while those with lower entropy (e.g., C1 with 0.82) see significantly reduced cache reuse at just 0.35. These results suggest that FedHLM effectively exploits semantic diversity for communication reduction, as diverse token generation leads to greater opportunities for cache hits or peer consensus. Furthermore, this analysis addresses potential reviewer concerns by showing that the framework adapts well to variation in token distribution complexity.

\begin{figure}[htbp]
    \centering
    \includegraphics[width=0.47\textwidth]{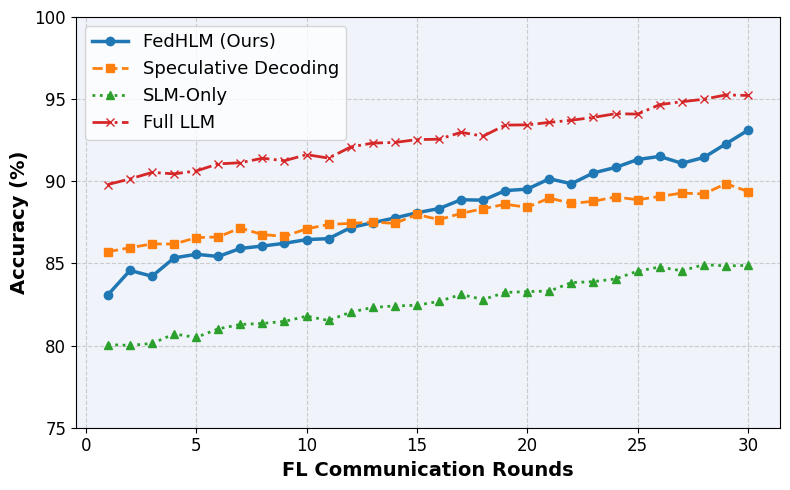} % Update path if needed
    \caption{Accuracy comparison of FedHLM with baseline methods over FL communication rounds.}
    \label{fig:accuracy_comparison}
\end{figure}

\begin{table}[htbp]
\centering
\caption{Inference Accuracy Comparison Between FedHLM and Centralized FL Over Selected FL Rounds}
\label{tab:accuracy_comparison}
\begin{tabular}{|c|c|c|}
\hline
\textbf{FL Round} & \textbf{FedHLM Accuracy (\%)} & \textbf{Centralized FL Accuracy (\%)} \\
\hline
1  & 83.2  & 82.46 \\
5  & 86.5  & 89.32 \\
10 & 88.3  & 93.73 \\
15 & 90.1  & 95.81 \\
20 & 91.4  & 96.79 \\
25 & 92.1  & 97.33 \\
30 & 93.2  & 97.67 \\
\hline
\end{tabular}
\end{table}

\subsection{Accuracy Comparison Over FL Rounds}

Fig.~\ref{fig:accuracy_comparison} shows the accuracy trends of FedHLM compared to three baseline methods: full LLM inference, speculative decoding, and SLM-only inference. As FL communication rounds progress, FedHLM steadily improves in accuracy, narrowing the gap with full LLM inference while significantly outperforming both speculative decoding and SLM-only approaches. This demonstrates the effectiveness of our FL-driven threshold adaptation and P2P model sharing in enabling high-quality inference with reduced reliance on cloud-based models.

FedHLM begins with lower accuracy but steadily improves over rounds, ultimately reaching an accuracy of 93.2\% at round 30. This trend illustrates the effectiveness of federated threshold adaptation and peer-to-peer model exchange, which enable continuous refinement of the routing policy and localized inference capabilities. Importantly, FedHLM reduces reliance on the cloud-hosted LLM while preserving most of its predictive power.

In comparison, speculative decoding shows modest improvements, plateauing at around 89.5\%, which suggests that its early-exit decisions are often suboptimal without collaboration or adaptive thresholding. The SLM-only baseline performs the weakest, remaining below 85\% even after 30 rounds. This confirms the limitations of using only lightweight local models without access to external knowledge or coordination.

Overall, these results demonstrate that FedHLM strikes a favorable balance between inference accuracy and communication efficiency. It leverages distributed learning to approach LLM-level performance while drastically reducing cloud interaction, making it particularly suitable for edge deployment in resource-constrained environments.

To further evaluate the benefits of our hierarchical FL design, Table~\ref{tab:accuracy_comparison} compares the inference accuracy of FedHLM with that of a centralized FL baseline across selected FL rounds. As expected, the centralized approach consistently achieves higher accuracy due to its global model aggregation, ultimately reaching 97.67\% by round~30. However, FedHLM provides highly competitive accuracy—achieving 93.2\% at round~30—while eliminating the need for a central server and significantly reducing communication overhead. This trade-off highlights the practicality of FedHLM in decentralized, resource-constrained environments where full synchronization is infeasible or undesirable.

%\begin{figure}[htbp]
    %\centering
    %\includegraphics[width=0.95\linewidth]{Pipeline.png}
    %\includegraphics[width=0.95\linewidth]{pics/Pipeline_b.png}
    %\caption{(a) FedHLM pipeline implemented using Kubeflow. Each component simulates a modular system function such as SLM inference, token routing, inter-cluster exchange, and LLM decision. (b) Zoomed-in view of the execution graph for detailed inspection of task dependencies and successful completion of each containerized function.}

    %\label{fig:kubeflow_pipeline}
%\end{figure}

\section{Conclusion and Future Work}
\label{sec:conclusion}

This paper has introduced FedHLM, a novel FL-enabled HLM framework that optimizes token-level communication decisions during inference. By integrating uncertainty-aware local inference, P2P exchange, and federated threshold adaptation, FedHLM enables efficient and scalable inference under edge resource constraints. This framework formalizes the token transmission process and proposes an adaptive learning mechanism to minimize uplink communication while maintaining inference accuracy. Experimental results demonstrate that FedHLM reduces LLM transmissions by over 95\% with negligible performance degradation, confirming the practicality of federated threshold adaptation for communication-efficient language modeling.

The proposed P2P resolution mechanism is particularly relevant in many real-world applications, such as predictive text input in regional languages with shared phrase structures, smart home or in-vehicle voice assistants with common utterances, and personalized question answering systems with overlapping domain-specific queries. Future work can further explore this direction by incorporating domain adaptation techniques or personalized clustering to improve semantic alignment and collaboration.

While the current implementation of FedHLM focuses on text-based token processing, the architecture is inherently compatible with cross-modal context adaptation. By leveraging dynamic token embeddings instead of modality-specific token IDs, the framework can be extended to support diverse modalities such as vision and speech. Future work may explore incorporating domain-specific clustering and dynamic importance weighting into federated threshold optimization to further improve adaptability in highly heterogeneous client scenarios. Additionally, expanding the FedHLM framework to multimodal inference scenarios, integrating unified semantic representations across vision, speech, and text modalities, could unlock substantial improvements in cross-modal federated edge computing applications.

Additional directions include personalized cache optimization, on-device fine-tuning of small language models, and adaptive client clustering strategies to further enhance scalability and personalization in real-world deployments.

%\bibliographystyle{cas-model2-names}

% Loading bibliography database
%\bibliography{cas-refs}
\bibliographystyle{IEEEtran}
\bibliography{cas-refs.bib}

% Generated by IEEEtran.bst, version: 1.14 (2015/08/26)
\begin{thebibliography}{10}
\providecommand{\url}[1]{#1}
\csname url@samestyle\endcsname
\providecommand{\newblock}{\relax}
\providecommand{\bibinfo}[2]{#2}
\providecommand{\BIBentrySTDinterwordspacing}{\spaceskip=0pt\relax}
\providecommand{\BIBentryALTinterwordstretchfactor}{4}
\providecommand{\BIBentryALTinterwordspacing}{\spaceskip=\fontdimen2\font plus
\BIBentryALTinterwordstretchfactor\fontdimen3\font minus \fontdimen4\font\relax}
\providecommand{\BIBforeignlanguage}[2]{{%
\expandafter\ifx\csname l@#1\endcsname\relax
\typeout{** WARNING: IEEEtran.bst: No hyphenation pattern has been}%
\typeout{** loaded for the language `#1'. Using the pattern for}%
\typeout{** the default language instead.}%
\else
\language=\csname l@#1\endcsname
\fi
#2}}
\providecommand{\BIBdecl}{\relax}
\BIBdecl

\bibitem{Solat2024_SFL}
F.~Solat, J.~Lee, and D.~Niyato, ``{Split Federated Learning-Empowered Energy-Efficient Mobile Traffic Prediction Over UAVs },'' \emph{IEEE Wireless Communications Letters}, vol.~13, no.~11, pp. 3064--3068, Nov. 2024.

\bibitem{Lee2024_FL}
J.~Lee, F.~Solat, T.~Y. Kim, and H.~V. Poor, ``{Federated Learning-Empowered Mobile Network Management for 5G and Beyond Networks: From Access to Core},'' \emph{IEEE Communications Surveys \& Tutorials}, vol.~26, no.~3, pp. 2176--2212, Jan. 2024.

\bibitem{shao2025division}
C.~Shao, X.~Hu, Y.~Lin, and F.~Xu, ``{Division-of-Thoughts: Harnessing Hybrid Language Model Synergy for Efficient On-Device Agents},'' arXiv: 2502.04392, 2025.

\bibitem{Solat2024_het}
F.~Solat, S.~Patni, S.~Lim, and J.~Lee, ``{Heterogeneous Privacy Level-Based Client Selection for Hybrid Federated and Centralized Learning in Mobile Edge Computing},'' \emph{IEEE Access}, vol.~12, pp. 108\,556--108\,572, July 2024.

\bibitem{qiao2025token}
L.~Qiao, M.~B. Mashhadi, Z.~Gao, R.~Tafazolli, M.~Bennis, and D.~Niyato, ``{Token Communications: A Unified Framework for Cross-modal Context-aware Semantic Communications},'' arXiv: 2502.12096, 2025.

\bibitem{Wang2025_uncertainity}
Z.~Wang, X.~Zhang, Z.~S. Li, and M.~Yan, ``{QoSBERT: An Uncertainty-Aware Approach based on Pre-trained Language Models for Service Quality Prediction},'' arXiv: 2505.07863, 2025.

\bibitem{Guo2024promptfl}
T.~Guo, S.~Guo, J.~Wang, X.~Tang, and W.~Xu, ``{PromptFL: Let Federated Participants Cooperatively Learn Prompts Instead of Models – Federated Learning in Age of Foundation Model},'' \emph{IEEE Transactions on Mobile Computing}, vol.~23, no.~5, pp. 5179--5194, May 2024.

\bibitem{mcmahan2017communication}
H.~B. McMahan, E.~Moore, D.~Ramage, S.~Hampson, and B.~A. y~Arcas, ``{Communication-Efficient Learning of Deep Networks from Decentralized Data},'' arXiv:1602.05629, 2017.

\bibitem{gu2023efficiently}
A.~Gu, K.~Goel, and C.~R{\'e}, ``{Efficiently Modeling Long Sequences with Structured State Spaces},'' arXiv:2111.00396, 2023.

\bibitem{chen2019speculative}
X.~L. Chen and M.~Bansal, ``{Speculative Decoding for Neural Sequence Models},'' in \emph{Proceedings of the 57th Annual Meeting of the Association for Computational Linguistics}, 2019, pp. 1393--1398.

\bibitem{Hao2024_HLM}
Z.~Hao, H.~Jiang, S.~Jiang, J.~Ren, and T.~Cao, ``{Hybrid SLM and LLM for Edge-Cloud Collaborative Inference},'' in \emph{Proceedings of the Workshop on Edge and Mobile Foundation Models}, 2024, pp. 36--41.

\bibitem{Hao2024_HLM_cite}
Q.~Hao, J.~Fan, F.~Xu, J.~Yuan, and Y.~Li, ``{HLM-Cite: Hybrid Language Model Workflow for Text-based Scientific Citation Prediction},'' arXiv: 2410.09112, 2025.

\bibitem{GarciaBernal2020}
D.~G. Bernal, ``Decentralizing large-scale natural language processing with federated learning,'' Master's Thesis, KTH Royal Institute of Technology, Stockholm, Sweden, 2020, degree Project in Computer Science and Engineering, Second Cycle, 30 Credits.

\bibitem{konecny2017federated}
J.~Konečný, H.~B. McMahan, F.~X. Yu, P.~Richtárik, A.~T. Suresh, and D.~Bacon, ``{Federated Learning: Strategies for Improving Communication Efficiency},'' in \emph{Proceedings of the 20th International Conference on Artificial Intelligence and Statistics}, 2017, pp. 306--314.

\bibitem{Yuan2022_dec}
B.~Yuan, Y.~He, J.~Davis, T.~Zhang, T.~Dao, B.~Chen, P.~Liang, C.~Ré, and C.~Zhang, ``{Decentralized Training of Foundation Models in Heterogeneous Environments},'' in \emph{Advances in Neural Information Processing Systems 35 (NeurIPS)}, vol.~35, 2022, pp. 25\,464--25\,477.

\bibitem{Lin2021_Fed_NLP}
B.~Y. Lin, C.~He, Z.~Zeng, H.~Wang, Y.~Huang, C.~Dupuy, R.~Gupta, M.~Soltanolkotabi, X.~Ren, and S.~Avestimehr, ``{FedNLP: Benchmarking Federated Learning Methods for Natural Language Processing Tasks},'' arXiv: 2104.08815, 2021.

\bibitem{Cai2023_FL}
D.~Cai, S.~Wang, Y.~Wu, F.~X. Lin, and M.~Xu, ``{Federated Few-Shot Learning for Mobile NLP},'' in \emph{Proceedings of 2023 the 29th Annual International Conference on Mobile Computing and Networking (ACM MobiCom)}, vol.~63, 2023, pp. 1--17.

\bibitem{Cai2023_NLP}
D.~Cai, Y.~Wu, S.~Wang, F.~X. Lin, and M.~Xu, ``{Efficient Federated Learning for Modern NLP},'' in \emph{Proceedings of the 2023 the 29th Annual International Conference on Mobile Computing and Networking (ACM MobiCom)}, vol.~37, 2023, pp. 1--16.

\bibitem{Liu2021_FL}
M.~Liu, S.~Ho, M.~Wang, L.~Gao, Y.~Jin, and H.~Zhang, ``{Federated Learning Meets Natural Language Processing: A Survey},'' arXiv: 2107.12603, 2021.

\bibitem{su2024titanic}
N.~Su, C.~Hu, B.~Li, and B.~Li, ``{TITANIC: Towards Production Federated Learning with Large Language Models},'' in \emph{Proceedings of the 2024 IEEE International Conference on Computer Communications (INFOCOM)}, 2024, pp. 611--620.

\bibitem{gal2016dropout}
Y.~Gal and Z.~Ghahramani, ``{Dropout as a Bayesian Approximation: Representing Model Uncertainty in Deep Learning},'' in \emph{Proceedings of the 33rd International Conference on Machine Learning}, 2016, pp. 1050--1059.

\bibitem{zhang2020uncertainty}
X.~Zhang, R.~Zhang, and S.~Cui, ``{Uncertainty-Aware Deep Learning Models for Wireless Communication Systems},'' \emph{IEEE Transactions on Communications}, vol.~68, no.~11, pp. 7172--7187, 2020.

\bibitem{Mobiny2021}
A.~Mobiny, P.~Yuan, S.~K. Moulik, N.~Garg, C.~C. Wu, and H.~V. Nguyen, ``{DropConnect is Effective in Modeling Uncertainty of Bayesian Deep Networks},'' \emph{Scientific Reports}, vol.~11, no.~1, 2021.

\bibitem{Labach2019}
A.~Labach, H.~Salehinejad, and S.~Valaee, ``{Survey of Dropout Methods for Deep Neural Networks},'' arXiv: 904.13310, 2019.

\bibitem{settles2009active}
B.~Settles, ``{Active Learning Literature Survey},'' \emph{University of Wisconsin, Madison}, vol.~52, no. 55-66, p.~11, 2009.

\bibitem{Yang2016_active}
Y.~Yang and M.~Loog, ``{Active Learning Using Uncertainty Information},'' in \emph{Proceedings of the 2016 the 23rd International Conference on Pattern Recognition (ICPR)}, Dec. 2016, pp. 2646--2651.

\bibitem{Zhu2010_active}
J.~Zhu, H.~Wang, B.~K. Tsou, and M.~Ma, ``{Active Learning With Sampling by Uncertainty and Density for Data Annotations},'' \emph{IEEE Transactions on Audio, Speech, and Language Processing}, vol.~18, no.~6, pp. 1323--1331, Sep. 2010.

\bibitem{Nguyen2022}
V.~L. Nguyen, M.~H. Shaker, and E.~Hüllermeier, ``{How to Measure Uncertainty in Uncertainty Sampling for Active Learning},'' \emph{Machine Learning}, vol. 111, no.~1, pp. 89--122, 2022.

\bibitem{blatz2004confidence}
J.~Blatz, E.~Fitzgerald, G.~Foster, C.~Goutte, A.~Kulesza, A.~Sanchis, and F.~J. Och, ``{Confidence Estimation for Natural Language Processing: A Study on Machine Translation},'' in \emph{Proceedings of the 20th International Conference on Computational Linguistics}, 2004, pp. 315--321.

\bibitem{Sharma2025}
A.~Sharma, H.~Devarajan, S.~Govindarajan, T.~B. Shanker, J.~A.~A. Rey, and S.~Nandi, ``{High-Confidence Classification of Partial Discharge Acoustic Signals Using Bayesian Networks for Uncertainty Quantification},'' \emph{IEEE Transactions on Instrumentation and Measurement}, vol.~74, p. 3506511, Jan. 2025.

\bibitem{Cheng_snapcfl}
Y.~Cheng, W.~Zhang, Z.~Zhang, J.~K. an~dQ. Xu, and S.~Wang, ``{SnapCFL: A Pre-clustering-based Clustered Federated Learning Framework for Data and System Heterogeneities},'' \emph{IEEE Transactions on Mobile Computing}, vol.~24, no.~6, pp. 5214--5228, June 2025.

\bibitem{li2019convergence}
X.~Li, K.~Huang, W.~Yang, S.~Wang, and Z.~Zhang, ``{On the Convergence of FedAvg on Non-IID Data},'' arXiv: 907.02189, 2019.

\bibitem{karimireddy2020scaffold}
S.~P. Karimireddy, S.~Kale, M.~Mohri, S.~Reddi, S.~Stich, and A.~T. Suresh, ``{SCAFFOLD: Stochastic Controlled Averaging for Federated Learning},'' in \emph{Proceedings of the 2020 the International conference on machine learning (PMLR)}, 2020, pp. 5132--5143.

\bibitem{khaled2020tighter}
A.~Khaled, K.~Mishchenko, and P.~Richtarik, ``{Tighter Theory for Local SGD on Identical and Heterogeneous Data},'' in \emph{Proceedings of the 2020 the International conference on Artificial Intelligence and Statistics (PMLR)}, 2020, pp. 4519--4529.

\bibitem{li2020fedprox}
T.~Li, A.~K. Sahu, M.~Zaheer, M.~Sanjabi, A.~Talwalkar, and V.~Smith, ``{Federated Optimization in Heterogeneous Networks},'' in \emph{Proceedings of the 2020 the International conference on Machine Learning and Systems 2 (MLSys)}, 2020, pp. 429--450.

\bibitem{wang2020fednova}
J.~Wang, Q.~Liu, H.~Liang, G.~Joshi, and H.~V. Poor, ``{Tackling the Objective Inconsistency Problem in Heterogeneous Federated Optimization},'' pp. 7611--7623, 2020.

\bibitem{zhang2015character}
X.~Zhang, J.~Zhao, and Y.~LeCun, ``{Character-level Convolutional Networks for Text Classification},'' in \emph{Advances in Neural Information Processing Systems (NeurIPS)}, 2015.

\bibitem{devlin2018bert}
J.~Devlin, M.~W. Chang, K.~Lee, and K.~Toutanova, ``{BERT: Pre-Training of Deep Bidirectional Transformers for Language Understanding},'' in \emph{Proceedings of the 2019 conference of the North American chapter of the association for computational linguistics: human language technologies}, June 2019, pp. 4171--4186.

\bibitem{wolf2020transformers}
T.~Wolf, L.~Debut, V.~Sanh, J.~Chaumond, C.~Delangue, A.~Moi, P.~Cistac, T.~Rault, R.~Louf, M.~Funtowicz, and J.~Brew, ``{Transformers: State-of-the-Art Natural Language Processing},'' in \emph{Proceedings of the 2020 conference on empirical methods in natural language processing: system demonstrations}, Oct. 2020, pp. 38--45.

\bibitem{hao2024hybrid}
Z.~Hao, H.~Jiang, J.~Ren, and T.~Cao, ``{Hybrid SLM and LLM for Edge-Cloud Collaborative Inference},'' in \emph{Proceedings of the 2024 the Workshop on Edge and Mobile Foundation Models}, 2024, pp. 36--41.

\end{thebibliography}
\begin{IEEEbiography}[{\includegraphics[width=1in,height=1.25in,clip,keepaspectratio]{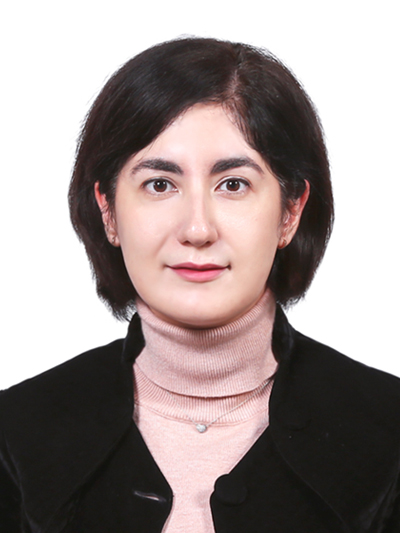}}]
{Faranaksadat Solat} received the B.S. degree in Industrial Engineering from Alzahra University and the M.S. degree in Information Technology Engineering from the Iran University of Science and Technology (IUST), Tehran, Iran, in 2017 and 2019, respectively. She obtained her Ph.D. degree in Software from the School of Computing, Gachon University, South Korea. She is currently an Assistant Professor in the School of Computing, where she conducts research in system optimization, machine learning (ML), deep learning (DL), federated learning (FL), 5G/6G network management, long short-term memory (LSTM), large language models (LLM), and machine learning operations (MLOps).
\end{IEEEbiography}

\begin{IEEEbiography}[{\includegraphics[width=1in,height=1.25in,clip,keepaspectratio]{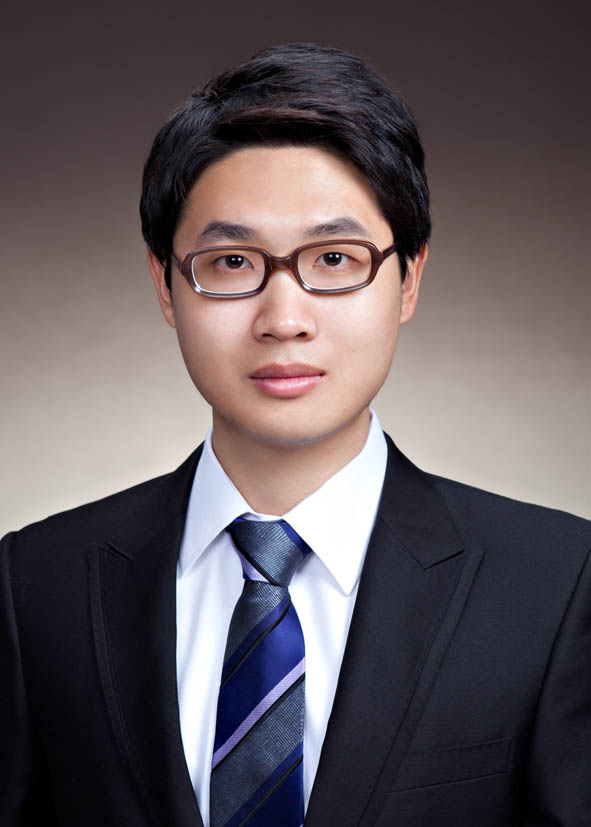}}]
{Joohyung Lee} received the B.S., M.S., and Ph.D. degrees from the Korea Advanced Institute of Science and Technology, Daejeon, South Korea, in 2008, 2010, and 2014, respectively. From 2012 to 2013, he was a Visiting Researcher with the Information Engineering Group, Department of Electronic Engineering, City University of Hong Kong, Hong Kong. From 2014 to 2017, he was a Senior Engineer with Samsung Electronics. He is currently an Associate Professor with the School of Computing, Gachon University, South Korea and also a Visiting Fellow with the Department of Electrical and Computer Engineering, Princeton University. He has contributed several articles to the International Telecommunication Union Telecommunication (ITU-T) and the 3rd Generation Partnership Project (3GPP). This current research interests include resource allocation, optimization and protocol design, with a focus on resource management for machine learning, including FL, 6G networks, cloud/edge computing, smart grids, augmented reality, virtual reality, and network economics. He received the Best Paper Award at the Integrated Communications, Navigation, and Surveillance Conference, in 2011, and Award for outstanding contribution in reviewing at Elsevier Computer Communications. He has been an IEEE Senior member since 2019, and a Technical Reviewer for several conferences and journals.
\end{IEEEbiography}

\begin{IEEEbiography}[{\includegraphics[width=1in,height=1.25in,clip,keepaspectratio]{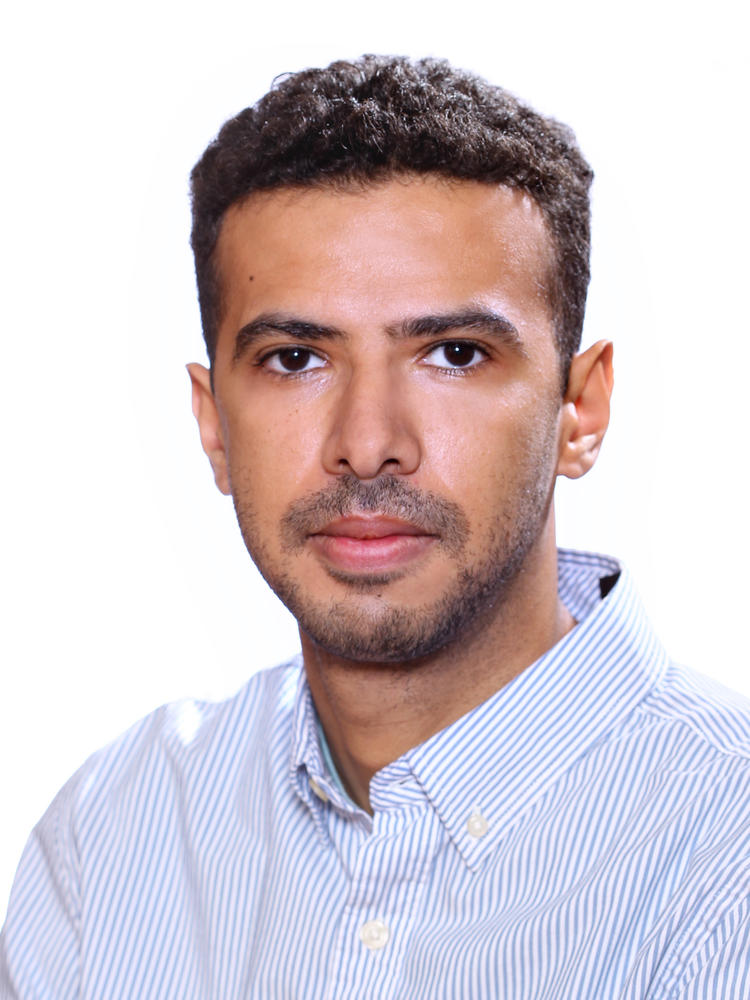}}]
{Mohamed Seif}
is currently a Post-Doctoral Research Associate with Princeton University. His current research interests include information theory, machine learning, and wireless communications.
\end{IEEEbiography}

\begin{IEEEbiography}[{\includegraphics[width=1in,height=1.25in,clip,keepaspectratio]{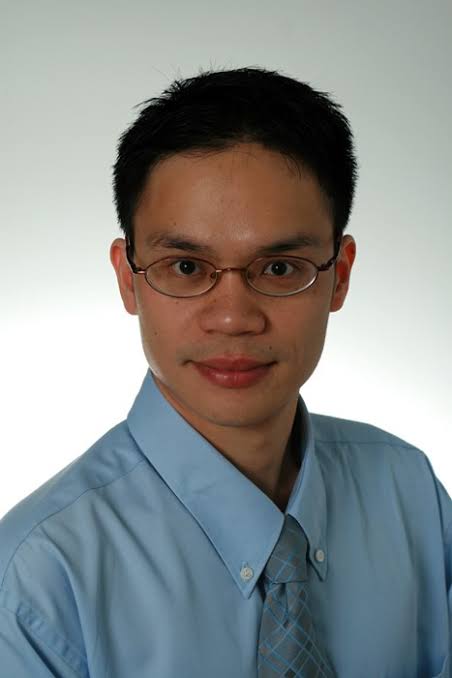}}]
{Dusit Niyato}
(Fellow, IEEE) is a professor in the School of Computer Science and Engineering, at Nanyang Technological University, Singapore. He received the B.Eng. degree from King Mongkut's Institute of Technology Ladkrabang (KMITL), Thailand, in 1999, and the Ph.D. degree in Electrical and Computer Engineering from the University of Manitoba, Canada, in 2008. His research interests include sustainability, edge intelligence, decentralized machine learning, and incentive mechanism design.
\end{IEEEbiography}

\begin{IEEEbiography}[{\includegraphics[width=1in,height=1.25in,clip,keepaspectratio]{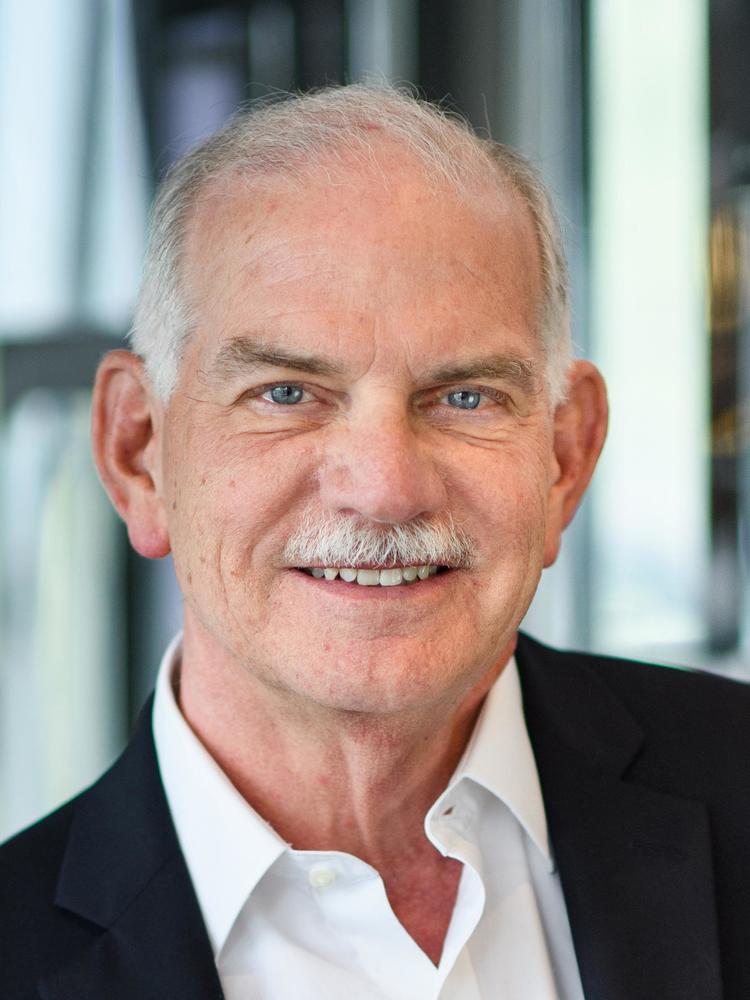}}]
{H. Vincent Poor} (S'72, M'77, SM'82, F'87) received the Ph.D. degree in EECS from Princeton University in 1977.  From 1977 until 1990, he was on the faculty of the University of Illinois at Urbana-Champaign. Since 1990 he has been on the faculty at Princeton, where he is currently the Michael Henry Strater University Professor. During 2006 to 2016, he served as the dean of Princeton’s School of Engineering and Applied Science. He has also held visiting appointments at several other universities, including most recently at Berkeley and Cambridge. His research interests are in the areas of information theory, machine learning and network science, and their applications in wireless networks, energy systems and related fields. Among his publications in these areas is the recent book \textit{Machine Learning and Wireless Communications}. (Cambridge University Press, 2022). Dr. Poor is a member of the National Academy of Engineering and the National Academy of Sciences and is a foreign member of the Chinese Academy of Sciences, the Royal Society, and other national and international academies. He received the IEEE Alexander Graham Bell Medal in 2017.
\end{IEEEbiography}

\end{document}